\documentclass[lettersize,journal]{IEEEtran}
\usepackage{amsmath,amsfonts}
\usepackage{algorithmic}
\usepackage{algorithm}
\usepackage{array}
\usepackage[caption=false,font=normalsize,labelfont=sf,textfont=sf]{subfig}
\usepackage{textcomp}
\usepackage{stfloats}
\usepackage{url}
\usepackage{verbatim}
\usepackage{graphicx}
\usepackage{cite}
\usepackage{booktabs}
\usepackage{amssymb}
\usepackage{multirow}
\newcommand{\inc}[1]{\,{\scriptsize $\uparrow$#1}}
\newcommand{\dec}[1]{\,{\scriptsize $\downarrow$#1}}
\newcommand{\tablefontsize}{\small}
\begin{document}

\title{Physics-Guided Spectral Distillation for Underwater Image Enhancement on Resource-Constrained Devices}

\author{Yifan Chen$^{\dagger}$, Kai He$^{\dagger}$, Ye Zheng, Jijun Lu, Zhe Sun$^{*}$, and Tao Chen$^{*}$,~\IEEEmembership{Senior Member,~IEEE}%
\thanks{Yifan Chen is with the College of Future Information Technology, Fudan University, Shanghai 200433, China, and also with the Institute of Artificial Intelligence (TeleAI), China Telecom, China (e-mail: chenyifan@fudan.edu.cn).}%
\thanks{Kai He is with the School of Computer Science and Technology, Harbin Institute of Technology, Weihai 264209, China (e-mail: 25B903170@stu.hit.edu.cn). Kai He conducted this work while interning at the Institute of Artificial Intelligence (TeleAI), China Telecom, China.}%
\thanks{Ye Zheng, Jijun Lu and Zhe Sun is with the Institute of Artificial Intelligence (TeleAI), China Telecom, China (e-mail: zhengye@westlake.edu.cn; jijun\_lu@163.com; sunzhe@nwpu.edu.cn.)}%

\thanks{Tao Chen is with the College of Future Information Technology, Fudan University, Shanghai 200433, China (e-mail: eetchen@fudan.edu.cn).}%
\thanks{$^{\dagger}$Yifan Chen and Kai He contributed equally to this work.}%
\thanks{$^{*}$Corresponding authors: Zhe Sun, and Tao Chen.}%
\thanks{This work was supported by the National Key R\&D Program of China under Grant 2026YFE0101200.}}

\markboth{}{}

\maketitle

\begin{abstract}
Underwater image enhancement is crucial for improving visual perception in marine applications. Existing underwater image enhancement studies mainly focus on enhancement quality and visual fidelity, while rarely considering real-time deployment capability, which is essential for resource-constrained underwater robots. To this end, we introduce a physics-guided spectral distillation (PSD) method, which reduces model capacity for real-time applications while maintaining the high performance of underwater image enhancement models. To decompose the outputs of teacher and student models, PSD adopts a multilevel Haar discrete wavelet transform. It transfers low-frequency color and illumination information as well as high-frequency structural details through band-specific objectives. Moreover, the distillation process of PSD is degradation-aware. We estimate degradation-aware weights through a physical head and combine them with ground-truth-guided reliability masks to selectively retain valuable teacher guidance. Experiments on the UIEB, LSUI, and EUVP datasets validate the effectiveness of the proposed method. Furthermore, we demonstrate the benefits of enhanced images for downstream perception tasks, including object detection. Deployment on a self-developed ROV further demonstrates its practical applicability in real-world underwater scenarios. 
\end{abstract}


\section{Introduction}
\label{sec:introduction}

\IEEEPARstart{U}{nderwater} image enhancement (UIE) improves image color, contrast, and detail to support underwater observation, robotic inspection, and object detection~\cite{sun2025wateroptical,sun2026extremedepth,jiang2024pdd}. However, water absorbs different wavelengths unequally and scatters light, leaving captured images with distorted colors, reduced contrast, and blurred details~\cite{jaffe1990model,akkaynak2018revised}. Correcting these coupled degradations remains challenging across real-world water and illumination conditions~\cite{zhou2024pdr,jiang2024pdd}. For underwater robots, enhancement must also keep pace with image acquisition under limited computation, memory, and power budgets. These resource constraints make preserving image quality at a low inference cost a central requirement for practical UIE~\cite{zhang2024liteenhancenet}.

\begin{figure}[!t]
    \centering
    \includegraphics[width=\columnwidth]{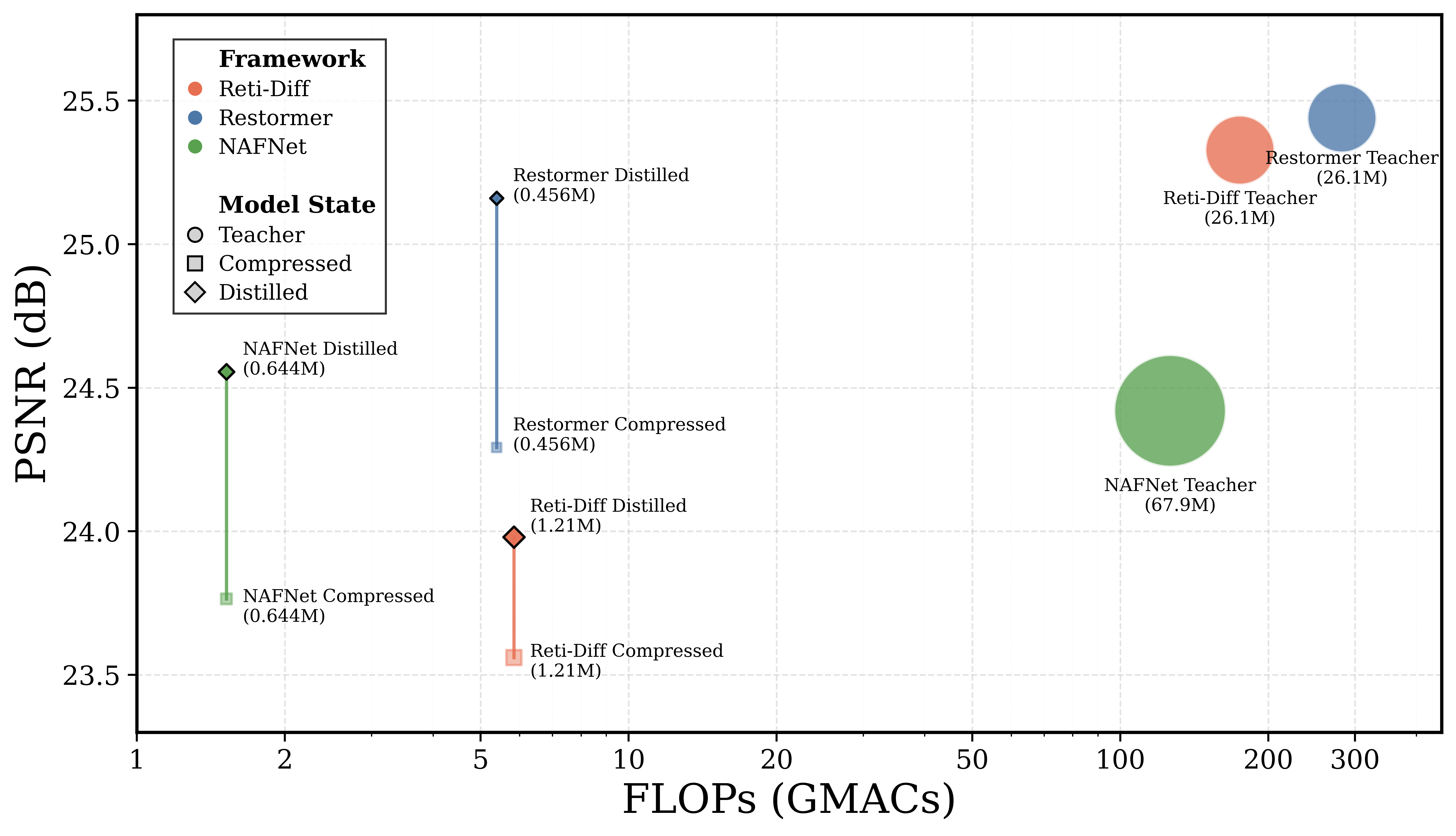}
    \caption{Accuracy--efficiency comparison across Reti-Diff, Restormer, and NAFNet. PSD improves each compact student at unchanged inference cost; circles denote teachers and annotations report parameter counts.}
    \label{fig:intro_efficiency}
\end{figure}

Traditional UIE methods use handcrafted enhancement rules or physical imaging models to correct underwater degradation. Multiscale fusion combines complementary image corrections to improve color and contrast~\cite{ancuti2012fusion}. Transmission priors and wavelength-aware models instead estimate light attenuation to recover scene appearance~\cite{drews2013transmission,akkaynak2018revised,akkaynak2019seathru}. Their effectiveness depends on whether the assumed image statistics or estimated optical parameters remain appropriate for the scene. Changes in water properties and illumination can therefore leave residual color casts or cause overcorrection. These limitations motivate learned restoration methods that accommodate a wider range of underwater conditions.

Learning-based UIE has advanced from convolutional restoration~\cite{li2020uwcnn} to Transformer, spatial--spectral, and diffusion architectures~\cite{peng2023ushape,peng2025ssuie,zhao2024wfdiff,he2025retidiff}. These models learn richer restoration mappings, but their computational and memory demands can limit real-time processing on embedded underwater platforms. Reducing network width and depth lowers inference cost, although the resulting capacity loss can weaken color recovery and boundary preservation. Knowledge distillation (KD) offers a way to improve compact students by transferring supervision from a high-capacity teacher during training~\cite{hinton2015distilling,xia2023mrda,zhou2025dckd}.

\begin{figure}[!t]
    \centering
    \includegraphics[width=\columnwidth]{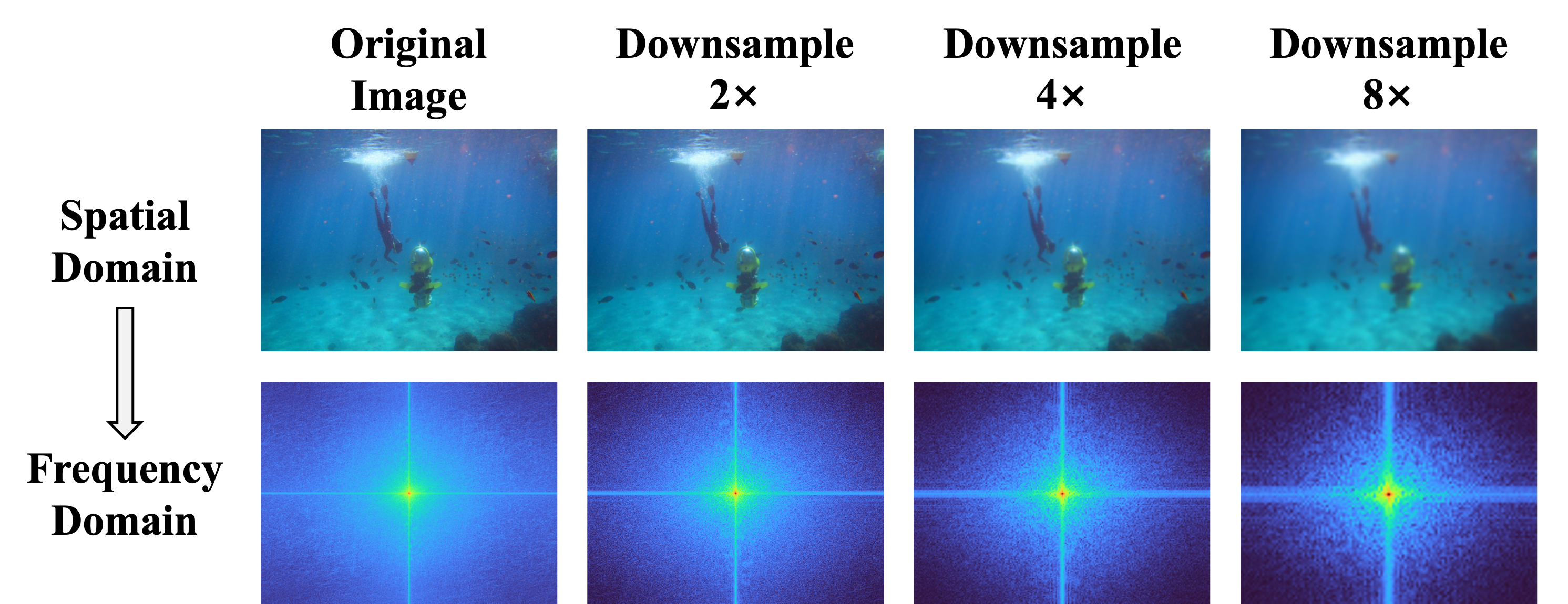}
    \caption{Spatial and spectral changes under downsampling by factors of $2$, $4$, and $8$, showing progressive high-frequency detail loss.}
    \label{fig:downsampling_frequency}
\end{figure}

Effective distillation for UIE requires selecting both what to transfer and where teacher guidance is useful. A single image-space matching objective does not separately control low-frequency appearance correction and high-frequency structural recovery. Frequency-aware KD provides separate spectral targets~\cite{zhang2022wkd,pham2024fam,zhang2024freekd}, but these targets alone do not describe spatially varying, channel-dependent underwater attenuation. Moreover, teacher predictions can remain inaccurate in particular regions, making indiscriminate imitation counterproductive. The downsampling examples in Fig.~\ref{fig:downsampling_frequency} illustrate how structural information can be weakened, motivating explicit attention to high-frequency details. Compact UIE students therefore need band-specific supervision that accounts for degradation severity and teacher reliability.

To address these requirements, we propose \emph{physics-guided spectral distillation} (PSD), an output-level distillation framework for compact UIE models. A multilevel Haar transform separates low-frequency color and illumination from high-frequency directional structures. Transmission-derived weights emphasize degraded regions and channels, while ground-truth-guided reliability masks retain spectral teacher targets only where they are more accurate than student predictions. An unmasked image-space loss complements spectral transfer by correcting residual RGB discrepancies. These objectives guide student training without retaining the teacher, physical head, or auxiliary computations at inference. Figure~\ref{fig:intro_efficiency} summarizes the resulting accuracy--efficiency improvements across Reti-Diff, Restormer, and NAFNet students.

The main contributions are summarized as follows:
\begin{itemize}
    \item We propose PSD, which combines multilevel band-specific spectral transfer and image-space compensation to preserve color, illumination, and structural information in compact UIE students.
    \item We introduce transmission-derived degradation weighting and teacher-reliability gating to focus distillation on degraded regions while filtering inaccurate teacher guidance, without inference-time overhead.
    \item Across three backbone--dataset settings, PSD enables compact students to reduce the teacher parameter counts by 95.45\%--99.06\%, while incurring average performance drops of only 0.501~dB in PSNR and 0.0059 in SSIM.
    \item We validate downstream utility on UDD and practical deployment on a self-developed remotely operated vehicle (ROV). Detection gains over raw inputs reach 3.22 and 1.85 percentage points in $\mathrm{mAP}_{50}$ and $\mathrm{mAP}_{50:95}$, respectively. The students achieve 35.7--84.6 FPS at $256\times256$ resolution.
\end{itemize}

\section{Related Work}
\label{sec:related_work}

\subsection{Underwater Image Enhancement}

UIE restores visibility, color, and structure degraded by underwater absorption and scattering. Classical models describe direct transmission and scattering~\cite{jaffe1990model}, while later work introduced wavelength-aware imaging models~\cite{akkaynak2018revised,akkaynak2019seathru}, transmission priors~\cite{drews2013transmission}, and multiscale fusion~\cite{ancuti2012fusion}. UIEB established a common real-image benchmark~\cite{li2020uieb}. Physics-aware learning incorporates transmission, color-space, reflectance, and diffusion priors~\cite{li2021ucolor,zhuang2022hyperlaplacian,li2026bridging}; PUGAN and GUPDM combine image formation with learned restoration~\cite{cong2023pugan,mu2023gupdm}.

Data-driven UIE ranges from adversarial and convolutional restoration~\cite{fabbri2018ugan,li2020uwcnn} to Transformers and reduced-supervision learning. TUDA addresses synthetic-to-real and intra-real domain gaps~\cite{wang2023tuda}, while U-Shape Transformer models long-range, spatially varying degradation~\cite{peng2023ushape}. Related computational ghost imaging methods use self-supervised information extraction and attention to improve underwater image reconstruction~\cite{chen2023iegi,chen2025aegi}. These advances motivate effective information recovery, while efficient enhancement of conventional RGB images remains the focus of PSD.

Frequency-domain and diffusion methods provide complementary priors. WF-Diff combines wavelet and Fourier representations~\cite{zhao2024wfdiff}; CPDM and Reti-Diff preserve content through diffusion and Retinex priors~\cite{shi2024cpdm,he2025retidiff}; and SS-UIE models non-uniform spatial--spectral degradation~\cite{peng2025ssuie}. Such designs improve global appearance and fine structure, but iterative or multi-branch processing increases memory and computation. Lightweight UIE networks consequently reduce parameters and operations for embedded underwater platforms~\cite{zhang2024liteenhancenet,zheng2024mfm}. These architecture-specific designs motivate a complementary question: whether a common distillation objective can compress different high-performance UIE backbones while preserving their restoration knowledge.

\subsection{Knowledge Distillation}

Knowledge distillation (KD) transfers teacher knowledge through softened outputs~\cite{hinton2015distilling}, intermediate features or attention~\cite{romero2015fitnets,zagoruyko2017attention}, and pairwise geometry~\cite{park2019rkd}. Curriculum-based distillation adjusts transfer difficulty during optimization~\cite{li2023ctkd}. These methods mainly target semantic representations rather than dense, frequency-sensitive restoration outputs.

Restoration-oriented KD transfers degradation and spectral knowledge. Wavelet KD emphasizes high-frequency signals~\cite{zhang2022wkd}, while MRDA distills implicit degradation representations~\cite{xia2023mrda}. Frequency Attention learns spectral filters~\cite{pham2024fam}, and FreeKD selects frequency components using semantic prompts and pixel-wise masks~\cite{zhang2024freekd}. DCKD adapts the distilled solution space to student learning~\cite{zhou2025dckd}. For UIE, an important distinction is therefore whether knowledge selection reflects the degradation of the input or only the informativeness of the teacher representation. However, these selection rules do not encode channel-dependent underwater attenuation. PSD couples transmission-derived weighting with ground-truth-guided reliability masks for spectral transfer, complemented by an unmasked RGB loss.

\section{Proposed Method}
\label{sec:method}

\subsection{Problem Formulation and Overview}
\label{subsec:formulation}

Let $\mathbf{I}\in[0,1]^{3\times H\times W}$ denote a degraded underwater image and $\mathbf{J}$ its paired reference. A frozen teacher $\mathcal{T}$ and a compact student $\mathcal{S}_{\theta}$ produce
\begin{equation}
    \mathbf{Y}^{t}=\mathcal{T}(\mathbf{I}), \qquad
    \mathbf{Y}^{s}=\mathcal{S}_{\theta}(\mathbf{I}),
    \label{eq:teacher_student}
\end{equation}
where $t$ and $s$ denote the teacher and student, respectively. We optimize $\mathcal{S}_{\theta}$ using its native restoration objective together with additional teacher supervision. The teacher and all PSD-specific components are used only during training. Only the compact student is deployed; PSD adds no inference-time modules.

\begin{figure*}[!t]
    \centering
    \includegraphics[width=\textwidth]{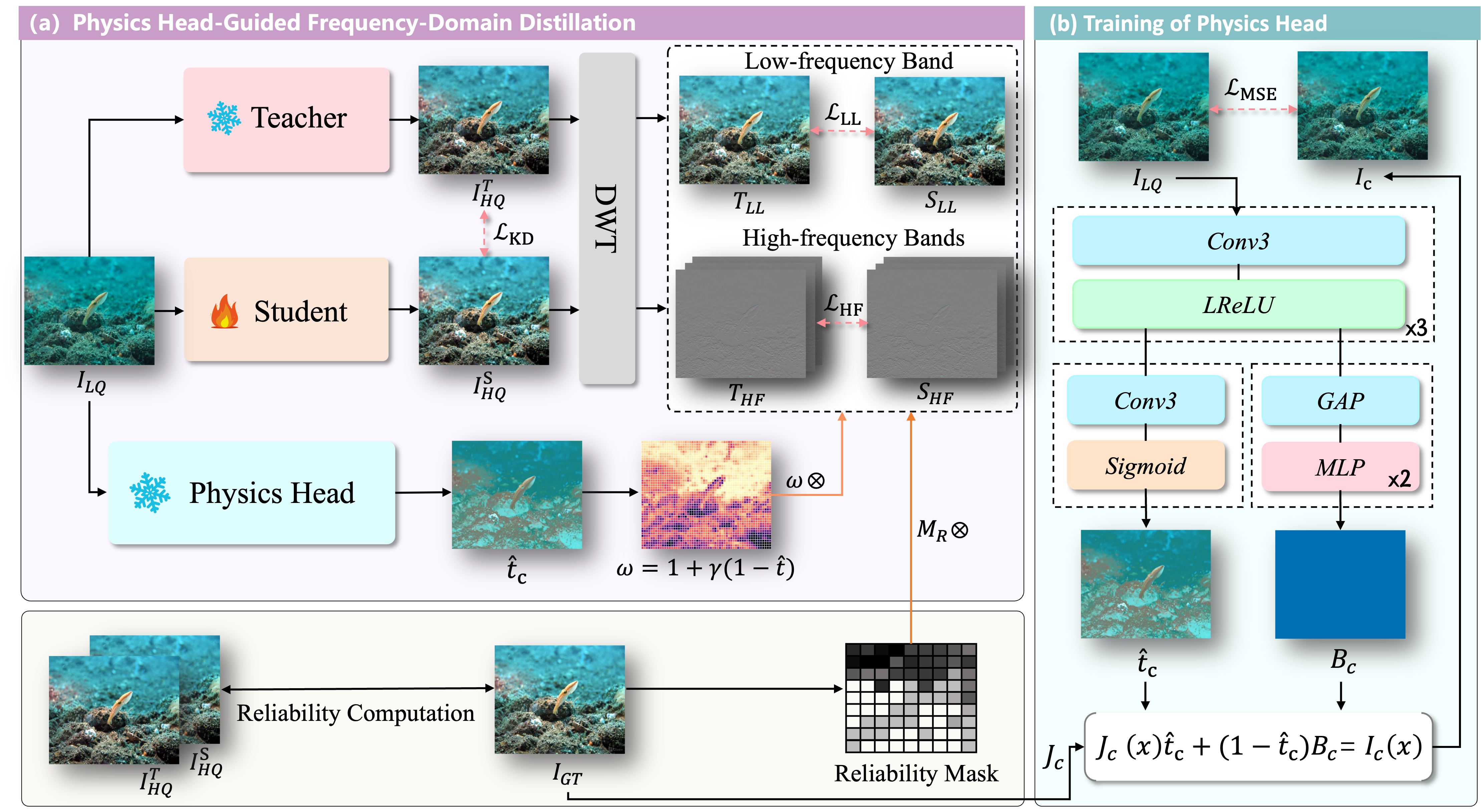}
    \caption{Overview of PSD. (a) The frozen teacher and compact student produce restored images that are decomposed into low- and high-frequency bands. Transmission-derived degradation weights $\boldsymbol{\omega}$ and ground-truth-guided reliability masks $\mathbf{M}_{R}$ regulate band-specific transfer, while the image-space term complements the spectral objectives. (b) The physical head is calibrated using the underwater image-formation model to estimate channel-wise transmission $\widehat{\mathbf{t}}$ and global background light $\widehat{\mathbf{B}}$. The teacher and all PSD-specific components are removed at inference.}
    \label{fig:method_overview}
\end{figure*}

Direct image matching does not distinguish spatially and spectrally varying degradation or unreliable teacher predictions. PSD estimates degradation with a physical head, decomposes restored outputs into wavelet bands, and gates spectral transfer by the teacher's advantage over the student. An unmasked RGB loss complements these objectives (Fig.~\ref{fig:method_overview}).

Algorithm~\ref{alg:psd_training} summarizes physical-head calibration followed by student training with task, spectral, and image-space losses. Here, $\eta_n$ is the auxiliary weight at training step $n$.

\begin{algorithm}[t]
\caption{Complete training procedure of PSD}
\label{alg:psd_training}
\footnotesize
\begin{algorithmic}[1]
\REQUIRE Paired training set $\mathcal{D}$; frozen teacher $\mathcal{T}$; student $\mathcal{S}_{\theta}$; physical head $\mathcal{H}_{\phi}$
\REQUIRE Native task loss $\mathcal{L}_{\mathrm{task}}$; wavelet depth $K$; auxiliary schedule $\eta_n$
\ENSURE Trained compact student $\mathcal{S}_{\theta}$
\STATE Initialize $\theta$ and $\phi$
\FOR{each physical-head calibration step}
    \STATE Sample $(\mathbf{I},\mathbf{J})\sim\mathcal{D}$
    \STATE $(\widehat{\mathbf{t}},\widehat{\mathbf{B}})\leftarrow\mathcal{H}_{\phi}(\mathbf{I})$
    \STATE Reconstruct $\widehat{\mathbf{I}}$ using \eqref{eq:re_degradation}
    \STATE Compute $\mathcal{L}_{\mathrm{phy}}$ using \eqref{eq:physics_loss}
    \IF{calibrating the Restormer or NAFNet implementation}
        \STATE $\mathcal{L}_{\mathrm{phy}}\leftarrow\mathcal{L}_{\mathrm{phy}}+0.1\|\nabla\widehat{\mathbf{I}}-\nabla\mathbf{I}\|_{1}$
    \ENDIF
    \STATE Update $\phi$ by minimizing $\mathcal{L}_{\mathrm{phy}}$
\ENDFOR
\STATE Freeze $\phi$
\FOR{each student-training step $n$}
    \STATE Sample $(\mathbf{I},\mathbf{J})\sim\mathcal{D}$
    \STATE $\mathbf{Y}^{t}\leftarrow\operatorname{sg}[\mathcal{T}(\mathbf{I})]$
    \STATE $\mathbf{Y}^{s}\leftarrow\mathcal{S}_{\theta}(\mathbf{I})$
    \STATE $(\widehat{\mathbf{t}},\widehat{\mathbf{B}})\leftarrow\operatorname{sg}[\mathcal{H}_{\phi}(\mathbf{I})]$
    \STATE Compute $\boldsymbol{\omega}$ from $\widehat{\mathbf{t}}$ using \eqref{eq:severity}--\eqref{eq:final_weight}
    \STATE Compute normalized $K$-level Haar bands using \eqref{eq:dwt_bands}--\eqref{eq:subband_normalization}
    \STATE Compute the LL and HF reliability masks using Algorithm~\ref{alg:reliability}
    \STATE Compute the spectral losses using \eqref{eq:ll_loss}--\eqref{eq:spectral_total}
    \STATE Compute the image-space and PSD losses using \eqref{eq:output_kd}--\eqref{eq:psd_total}
    \STATE $\mathcal{L}_{\mathrm{total}}\leftarrow\mathcal{L}_{\mathrm{task}}(\mathbf{Y}^{s},\mathbf{J})+\eta_n\mathcal{L}_{\mathrm{PSD}}$
    \STATE Update $\theta$ by minimizing $\mathcal{L}_{\mathrm{total}}$
\ENDFOR
\STATE Remove $\mathcal{T}$ and $\mathcal{H}_{\phi}$; return $\mathcal{S}_{\theta}$
\end{algorithmic}
\end{algorithm}

\subsection{Underwater-Formation-Guided Degradation Weighting}
\label{subsec:physics}

\subsubsection{Physical-head calibration}

We adopt the channel-dependent underwater image-formation model
\begin{equation}
    \mathbf{I}_{c}(\mathbf{x})
    =
    \mathbf{J}_{c}(\mathbf{x})\mathbf{t}_{c}(\mathbf{x})
    +
    \mathbf{B}_{c}\left[1-\mathbf{t}_{c}(\mathbf{x})\right],
    \label{eq:uifm}
\end{equation}
where $c\in\{r,g,b\}$ denotes a color channel and $\mathbf{x}$ indexes a spatial location. Here, $\mathbf{t}$ is the channel-wise transmission, and $\mathbf{B}$ is the global background light. A lightweight physical head $\mathcal{H}_{\phi}$ predicts both quantities from the degraded input. Its shared trunk comprises three $3\times3$ convolutional layers with 32 channels and Leaky-ReLU activations. Given the trunk feature $\mathbf{F}=\mathcal{H}_{\phi}^{\mathrm{trunk}}(\mathbf{I})$, the two prediction branches are
\begin{align}
    \widehat{\mathbf{t}}
    &=
    t_{\min}+(t_{\max}-t_{\min})
    \sigma\!\left(\operatorname{Conv}_{t}(\mathbf{F})\right),
    \label{eq:transmission_head}\\
    \widehat{\mathbf{B}}
    &=
    \sigma\!\left(
    \operatorname{MLP}_{B}(\operatorname{GAP}(\mathbf{F}))
    \right).
    \label{eq:background_head}
\end{align}
We set $t_{\min}=0.05$ and $t_{\max}=0.99$ to bound the decomposition.

We calibrate the physical head before using it for distillation. The degraded observation is reconstructed as
\begin{equation}
    \widehat{\mathbf{I}}
    =
    \mathbf{J}\odot\widehat{\mathbf{t}}
    +
    \widehat{\mathbf{B}}\odot
    (1-\widehat{\mathbf{t}}),
    \label{eq:re_degradation}
\end{equation}
where $\widehat{\mathbf{B}}$ is broadcast spatially. The calibration objective is
\begin{equation}
    \mathcal{L}_{\mathrm{phy}}
    =
    \left\|\widehat{\mathbf{I}}-\mathbf{I}\right\|_{1}
    +
    \lambda_{\mathrm{tv}}\operatorname{TV}
    (\widehat{\mathbf{t}}).
    \label{eq:physics_loss}
\end{equation}
We set $\lambda_{\mathrm{tv}}=0.05$. During this phase, the reference image is detached so that only the physical head is updated. For the standalone Restormer and NAFNet implementations, calibration also includes $0.1\|\nabla\widehat{\mathbf{I}}-\nabla\mathbf{I}\|_{1}$. This auxiliary term stabilizes the physical estimator but does not supervise the student. After calibration, $\mathcal{H}_{\phi}$ remains frozen.

\subsubsection{Transmission-derived degradation weight}

Low transmission indicates strong attenuation. We therefore convert the calibrated transmission into a channel-wise spatial weight
\begin{align}
    \widetilde{\boldsymbol{\omega}}
    &=
    1+\gamma(1-\widehat{\mathbf{t}}),
    \label{eq:severity}\\
    \boldsymbol{\omega}
    &=
    \frac{\widetilde{\boldsymbol{\omega}}}
    {\operatorname{mean}
    (\widetilde{\boldsymbol{\omega}})+\epsilon}.
    \label{eq:final_weight}
\end{align}
This normalization fixes the mean scale of the resulting weight map. We set $\gamma=2$ and detach the weight before applying the distillation losses. Consequently, the physical head cannot alter its transmission prediction merely to reduce the teacher--student discrepancy. Background light supports calibration of the image-formation model, whereas transmission provides the spatial and channel weighting used by PSD.

\subsection{Reliability-Aware Multilevel Spectral Distillation}
\label{subsec:spectral}

\subsubsection{Two-level Haar representation}

Following the multiresolution wavelet formulation~\cite{mallat1989wavelet}, PSD applies the discrete wavelet transform directly to restored outputs rather than intermediate features or residuals. For a $2\times2$ neighborhood $(a,b,c,d)$, the channel-preserving Haar transform is
\begin{equation}
\begin{aligned}
    L &= (a+b+c+d)/2, &
    H^{h} &= (-a-b+c+d)/2,\\
    H^{v} &= (-a+b-c+d)/2, &
    H^{d} &= (a-b-c+d)/2.
\end{aligned}
\label{eq:haar}
\end{equation}
The low-frequency band $L$ primarily represents coarse intensity and chromatic content. The signed bands $H^{h}$, $H^{v}$, and $H^{d}$ represent horizontal, vertical, and diagonal structures.

Let $q\in\{s,t,g\}$ index the student, teacher, and ground truth, respectively. We set $\mathbf{L}_{0}^{q}=\mathbf{Y}^{q}$ and $\mathbf{Y}^{g}=\mathbf{J}$. At level $\ell$, we recursively compute
\begin{equation}
    (\mathbf{L}_{\ell}^{q},
     \mathbf{H}_{\ell,h}^{q},
     \mathbf{H}_{\ell,v}^{q},
     \mathbf{H}_{\ell,d}^{q})
    =
    \mathcal{W}(\mathbf{L}_{\ell-1}^{q}).
    \label{eq:dwt_bands}
\end{equation}
We use $K=2$ decomposition levels. To make coefficient magnitudes comparable across resolutions, we define
\begin{equation}
    \overline{\mathbf{L}}_{K}^{q}
    =\mathbf{L}_{K}^{q}/2^{K},
    \qquad
    \overline{\mathbf{H}}_{\ell,b}^{q}
    =\mathbf{H}_{\ell,b}^{q}/2^{\ell},
    \label{eq:subband_normalization}
\end{equation}
where $b\in\{h,v,d\}$. Before each transform, we crop odd spatial dimensions by one pixel.

\subsubsection{Ground-truth-guided teacher-advantage masks}

Uniform teacher imitation may propagate coefficients that are less accurate than those already produced by the student. PSD therefore compares detached teacher and student errors with the paired reference. For the final low-frequency band, we compute
\begin{equation}
    \mathbf{e}_{\mathrm{LL}}^{q}
    =
    (\overline{\mathbf{L}}_{K}^{q}
    -\overline{\mathbf{L}}_{K}^{g})^{2},
    \qquad q\in\{s,t\}.
    \label{eq:ll_reliability_error}
\end{equation}
For the three high-frequency orientations at level $\ell$, the channel-wise error is
\begin{equation}
    \mathbf{e}_{\mathrm{HF},\ell}^{q}
    =
    \frac{1}{3}
    \sum_{b\in\{h,v,d\}}
    (\overline{\mathbf{H}}_{\ell,b}^{q}
    -\overline{\mathbf{H}}_{\ell,b}^{g})^{2}.
    \label{eq:hf_reliability_error}
\end{equation}
The corresponding hard masks are
\begin{align}
    \mathbf{M}_{\mathrm{LL}}
    &=
    \operatorname{sg}\!\left[
    \mathbf{1}(\mathbf{e}_{\mathrm{LL}}^{t}
    <\mathbf{e}_{\mathrm{LL}}^{s})\right],
    \label{eq:ll_reliability_mask}\\
    \mathbf{M}_{\mathrm{HF},\ell}
    &=
    \operatorname{sg}\!\left[
    \mathbf{1}(\mathbf{e}_{\mathrm{HF},\ell}^{t}
    <\mathbf{e}_{\mathrm{HF},\ell}^{s})\right],
    \label{eq:hf_reliability_mask}
\end{align}
where $\operatorname{sg}$ denotes the stop-gradient operation. A coefficient therefore contributes to knowledge transfer only when the teacher is closer to the reference than the current student. The masks are recomputed online as the student evolves.

\subsubsection{Band-specific spectral objectives}

We downsample $\boldsymbol{\omega}$ to match the resolution of each subband. The final LL band retains the channel-wise weight $\boldsymbol{\omega}_{K}$. High-frequency transfer instead uses the channel average $\overline{\boldsymbol{\omega}}_{\ell}$. The low-frequency objective is
\begin{equation}
    \mathcal{L}_{\mathrm{LL}}
    =
    \mathbb{E}\!\left[
    \alpha_{\mathrm{LL}}
    \boldsymbol{\omega}_{K}
    \odot\mathbf{M}_{\mathrm{LL}}
    \odot
    (\overline{\mathbf{L}}_{K}^{s}
    -\overline{\mathbf{L}}_{K}^{t})^{2}
    \right].
    \label{eq:ll_loss}
\end{equation}
This term transfers coarse intensity and channel-dependent chromatic correction. The high-frequency objective is
\begin{equation}
\begin{split}
    \mathcal{L}_{\mathrm{HF}}
    ={}&
    \frac{1}{\sum_{\ell=1}^{K}\rho^{\ell-1}}
    \sum_{\ell=1}^{K}\rho^{\ell-1}
    \mathbb{E}\Bigg[
    \alpha_{\mathrm{HF}}
    \overline{\boldsymbol{\omega}}_{\ell}
    \odot\mathbf{M}_{\mathrm{HF},\ell}\\
    &\qquad\qquad\odot
    \sum_{b\in\{h,v,d\}}
    (\overline{\mathbf{H}}_{\ell,b}^{s}
    -\overline{\mathbf{H}}_{\ell,b}^{t})^{2}
    \Bigg],
    \label{eq:hf_loss}
\end{split}
\end{equation}
where $\rho=0.5$ reduces the contribution of coarser high-frequency levels. We set the band factors to $\alpha_{\mathrm{LL}}=1$ and $\alpha_{\mathrm{HF}}=0.5$. The spectral loss is
\begin{equation}
    \mathcal{L}_{\mathrm{spec}}
    =
    \lambda_{\mathrm{LL}}\mathcal{L}_{\mathrm{LL}}
    +
    \lambda_{\mathrm{HF}}\mathcal{L}_{\mathrm{HF}},
    \label{eq:spectral_total}
\end{equation}
with $\lambda_{\mathrm{LL}}=1$ and $\lambda_{\mathrm{HF}}=0.1$.

\subsection{Image-Space Compensation}
\label{subsec:output_kd}

The wavelet objectives separate appearance and structural supervision. We additionally use a low-weight image-space loss to align the student and teacher RGB outputs. In our experiments, reliability gating was applied only to the LL and HF spectral losses; the image-space loss used no reliability mask.

Algorithm~\ref{alg:reliability} summarizes the reliability computation for the spectral objectives, preserving the native channel structure of the corresponding coefficients.

\begin{algorithm}[t]
\caption{Ground-truth-guided reliability computation}
\label{alg:reliability}
\footnotesize
\begin{algorithmic}[1]
\REQUIRE Normalized bands $\{\overline{\mathbf{L}}_{K}^{q},\overline{\mathbf{H}}_{\ell,b}^{q}\}$ for $q\in\{s,t,g\}$
\ENSURE Reliability masks for the LL and HF spectral bands
\FOR{$q\in\{s,t\}$}
    \STATE $\mathbf{e}_{\mathrm{LL}}^{q}\leftarrow(\overline{\mathbf{L}}_{K}^{q}-\overline{\mathbf{L}}_{K}^{g})^{2}$
\ENDFOR
\STATE $\mathbf{M}_{\mathrm{LL}}\leftarrow\operatorname{sg}[\mathbf{1}(\mathbf{e}_{\mathrm{LL}}^{t}<\mathbf{e}_{\mathrm{LL}}^{s})]$
\FOR{$\ell=1$ to $K$}
    \FOR{$q\in\{s,t\}$}
        \STATE Compute $\mathbf{e}_{\mathrm{HF},\ell}^{q}$ using \eqref{eq:hf_reliability_error}
    \ENDFOR
    \STATE $\mathbf{M}_{\mathrm{HF},\ell}\leftarrow\operatorname{sg}[\mathbf{1}(\mathbf{e}_{\mathrm{HF},\ell}^{t}<\mathbf{e}_{\mathrm{HF},\ell}^{s})]$
\ENDFOR
\STATE \textbf{return} $\mathbf{M}_{\mathrm{LL}}$, $\{\mathbf{M}_{\mathrm{HF},\ell}\}_{\ell=1}^{K}$
\end{algorithmic}
\end{algorithm}

We then define
\begin{equation}
    \mathcal{L}_{\mathrm{img}}
    =
    \mathbb{E}\!\left[
    (\mathbf{Y}^{s}-\operatorname{sg}[\mathbf{Y}^{t}])^{2}
    \right].
    \label{eq:output_kd}
\end{equation}
The expectation averages squared differences over all RGB channels, pixels, and batch samples, with the teacher output detached from gradient computation. This term provides global output alignment alongside the selectively gated spectral supervision.

The complete distillation objective is
\begin{equation}
    \mathcal{L}_{\mathrm{PSD}}
    =
    \mathcal{L}_{\mathrm{spec}}
    +
    \lambda_{\mathrm{img}}\mathcal{L}_{\mathrm{img}},
    \qquad \lambda_{\mathrm{img}}=0.25.
    \label{eq:psd_total}
\end{equation}

\section{Experiments and Application Validation}
\label{sec:experiments}

\subsection{Experimental Setup}
\label{subsec:experimental_setup}

\subsubsection{Datasets and evaluation metrics}

We used the 890 reference-paired images from UIEB's 950 images~\cite{li2020uieb}. LSUI provides 4,279 pairs covering scenes, water types, and illumination conditions~\cite{peng2023ushape}. For EUVP~\cite{islam2020fast}, we merged EUVP$-$I, EUVP$-$D, and EUVP$-$S, and divided the combined data into training and validation sets at an 8:2 ratio. All teachers, students, and distillation baselines shared identical partitions within each backbone--dataset setting. We report peak signal-to-noise ratio (PSNR) and structural similarity (SSIM), computed against paired references on RGB outputs without border cropping. Scores were averaged over each evaluation split, using the validation set for EUVP; higher values indicate better fidelity. All experiments used one fixed random seed, without repeated-run confidence intervals.

\subsubsection{Compared distillation methods}

For each dataset, the original Reti-Diff~\cite{he2025retidiff}, Restormer~\cite{zamir2022restormer}, or NAFNet~\cite{chen2022nafnet} model served as the frozen teacher. We compare six settings: the teacher, the task-only compact student, CTKD~\cite{li2023ctkd}, DCKD~\cite{zhou2025dckd}, FreeKD+~\cite{zhang2026freekdplus}, and PSD. Within each backbone--dataset setting, all distillation methods used the same teacher checkpoint, student architecture, data split, task loss, and training budget. These controls allow comparisons among student variants at a fixed model capacity.

\begin{table*}[t]
\centering
\caption{Complexity of the teacher and compressed student models for a $256\times256$ input. Retention is the student-to-teacher ratio.}
\label{tab:complexity}
\tablefontsize
\setlength{\tabcolsep}{4.5pt}
\begin{tabular*}{0.96\textwidth}{@{\extracolsep{\fill}}llcccccc@{}}
\toprule
Backbone & Source & \multicolumn{2}{c}{Parameters (M)} & \multicolumn{2}{c}{GMACs} & Param. retained & GMACs retained \\
\cmidrule(lr){3-4}\cmidrule(lr){5-6}
 & & Teacher & Student & Teacher & Student & (\%) & (\%) \\
\midrule
Reti-Diff~\cite{he2025retidiff} & ICLR' 2025 & 26.13 & 1.19 & 175.01 & 4.32 & 4.6 & 2.5 \\
Restormer~\cite{zamir2022restormer} & CVPR' 2022 & 26.12 & 0.46 & 281.98 & 5.39 & 1.8 & 1.9 \\
NAFNet~\cite{chen2022nafnet} & ECCV' 2022 & 67.89 & 0.64 & 126.17 & 1.52 & 0.94 & 1.2 \\
\bottomrule
\end{tabular*}
\end{table*}


\begin{table*}[t]
\centering
\caption{Quantitative comparison on UIEB, LSUI, and the merged EUVP dataset across three backbone architectures. 
Bold and underlined values denote the best and second-best results among compact student variants within each backbone, respectively. 
Teacher models are references; arrows show changes from the task-only student.}
\label{tab:main_results}

\tablefontsize
\setlength{\tabcolsep}{2.5pt}
\renewcommand{\arraystretch}{1.12}

\begin{tabular*}{0.96\textwidth}{@{\extracolsep{\fill}}clcccccc@{}}
\toprule
\multirow{2}{*}{\textbf{Reti-Diff}}
& \multirow{2}{*}{Source}
& \multicolumn{2}{c}{UIEB}
& \multicolumn{2}{c}{LSUI}
& \multicolumn{2}{c}{Merged EUVP} \\
\cmidrule(lr){3-4}\cmidrule(lr){5-6}\cmidrule(lr){7-8}
& & PSNR $\uparrow$ & SSIM $\uparrow$
& PSNR $\uparrow$ & SSIM $\uparrow$
& PSNR $\uparrow$ & SSIM $\uparrow$ \\
\midrule
Teacher & --
& 24.54 & 0.9341
& 28.59 & 0.8782
& 22.86 & 0.8191 \\

Task only & --
& 24.18 & 0.9263
& \underline{24.39} & 0.8680
& 22.12 & 0.8166 \\

CTKD~\cite{li2023ctkd} & AAAI' 2023
& 23.76\dec{0.42} & 0.9239\dec{0.0024}
& 23.40\dec{0.99} & 0.8629\dec{0.0051}
& 22.16\inc{0.04} & 0.8202\inc{0.0036} \\

DCKD~\cite{zhou2025dckd} & AAAI' 2025
& \underline{24.72}\inc{0.54} & \textbf{0.9324}\inc{0.0061}
& 22.13\dec{2.26} & 0.8474\dec{0.0206}
& \textbf{22.57}\inc{0.45} & \underline{0.8218}\inc{0.0052} \\

FreeKD+~\cite{zhang2026freekdplus} & TPAMI' 2026
& 23.74\dec{0.44} & 0.9236\dec{0.0027}
& 23.06\dec{1.33} & \underline{0.8687}\inc{0.0007}
& 22.28\inc{0.16} & 0.8156\dec{0.0010} \\

PSD (ours) & --
& \textbf{24.77}\inc{0.59} & \underline{0.9301}\inc{0.0038}
& \textbf{24.61}\inc{0.22} & \textbf{0.8694}\inc{0.0014}
& \underline{22.56}\inc{0.44} & \textbf{0.8227}\inc{0.0061} \\
\bottomrule
\end{tabular*}

\vspace{4pt}

\begin{tabular*}{0.96\textwidth}{@{\extracolsep{\fill}}clcccccc@{}}
\toprule
\multirow{2}{*}{\textbf{Restormer}}
& \multirow{2}{*}{Source}
& \multicolumn{2}{c}{UIEB}
& \multicolumn{2}{c}{LSUI}
& \multicolumn{2}{c}{Merged EUVP} \\
\cmidrule(lr){3-4}\cmidrule(lr){5-6}\cmidrule(lr){7-8}
& & PSNR $\uparrow$ & SSIM $\uparrow$
& PSNR $\uparrow$ & SSIM $\uparrow$
& PSNR $\uparrow$ & SSIM $\uparrow$ \\
\midrule
Teacher & --
& 23.01 & 0.9204
& 28.97 & 0.9134
& 24.35 & 0.8844 \\

Task only & --
& 22.98 & 0.9136
& 27.79 & 0.8964
& 22.12 & 0.8166 \\

CTKD~\cite{li2023ctkd} & AAAI' 2023
& \underline{23.23}\inc{0.25} & \textbf{0.9188}\inc{0.0052}
& \underline{28.17}\inc{0.38} & \textbf{0.9014}\inc{0.0050}
& 22.16\inc{0.04} & 0.8202\inc{0.0036} \\

DCKD~\cite{zhou2025dckd} & AAAI' 2025
& 23.09\inc{0.11} & 0.9169\inc{0.0033}
& 27.92\inc{0.13} & \underline{0.9002}\inc{0.0038}
& \textbf{23.96}\inc{1.84} & \textbf{0.8684}\inc{0.0518} \\

FreeKD+~\cite{zhang2026freekdplus} & TPAMI' 2026
& 22.79\dec{0.19} & 0.9093\dec{0.0043}
& 27.84\inc{0.05} & 0.8969\inc{0.0005}
& 23.86\inc{1.74} & 0.8664\inc{0.0498} \\

PSD (ours) & --
& \textbf{23.34}\inc{0.36} & \underline{0.9185}\inc{0.0049}
& \textbf{28.18}\inc{0.39} & 0.8996\inc{0.0032}
& \underline{23.95}\inc{1.83} & \underline{0.8677}\inc{0.0511} \\
\bottomrule
\end{tabular*}

\vspace{4pt}

\begin{tabular*}{0.96\textwidth}{@{\extracolsep{\fill}}clcccccc@{}}
\toprule
\multirow{2}{*}{\textbf{NAFNet}}
& \multirow{2}{*}{Source}
& \multicolumn{2}{c}{UIEB}
& \multicolumn{2}{c}{LSUI}
& \multicolumn{2}{c}{Merged EUVP} \\
\cmidrule(lr){3-4}\cmidrule(lr){5-6}\cmidrule(lr){7-8}
& & PSNR $\uparrow$ & SSIM $\uparrow$
& PSNR $\uparrow$ & SSIM $\uparrow$
& PSNR $\uparrow$ & SSIM $\uparrow$ \\
\midrule
Teacher & --
& 22.20 & 0.8969
& 26.84 & 0.8971
& 24.22 & 0.8752 \\

Task only & --
& 22.12 & \underline{0.9002}
& 25.31 & 0.8757
& 23.85 & 0.8651 \\

CTKD~\cite{li2023ctkd} & AAAI' 2023
& 22.24\inc{0.12} & 0.8987\dec{0.0015}
& \underline{25.73}\inc{0.42} & \underline{0.8778}\inc{0.0021}
& 23.88\inc{0.03} & 0.8650\dec{0.0001} \\

DCKD~\cite{zhou2025dckd} & AAAI' 2025
& 22.24\inc{0.12} & 0.8961\dec{0.0041}
& 25.54\inc{0.23} & 0.8746\dec{0.0011}
& \underline{23.90}\inc{0.05} & \underline{0.8658}\inc{0.0007} \\

FreeKD+~\cite{zhang2026freekdplus} & TPAMI' 2026
& \underline{22.34}\inc{0.22} & 0.8966\dec{0.0036}
& 25.54\inc{0.23} & 0.8766\inc{0.0009}
& 23.89\inc{0.04} & 0.8649\dec{0.0002} \\

PSD (ours) & --
& \textbf{22.47}\inc{0.35} & \textbf{0.9027}\inc{0.0025}
& \textbf{27.28}\inc{1.97} & \textbf{0.8889}\inc{0.0132}
& \textbf{23.91}\inc{0.06} & \textbf{0.8661}\inc{0.0010} \\
\bottomrule
\end{tabular*}

\end{table*}

\subsubsection{Cross-backbone students and training protocol}

The students cover diffusion-guided, Transformer, and convolutional restoration. Compact Reti-Diff retains the original dual-prior interface and uses a 128-dimensional prior, one deterministic reverse step, base width 16, and one block per level. Restormer and NAFNet use base width 12 and one block per encoder--decoder stage. Each student followed the training strategy and native task objective of its original framework. For PSD, we calibrated and froze the physical head before applying the two-level spectral and image-space losses in Sec.~\ref{sec:method}. Because PSD operates on restored outputs, it requires no feature adapters or matched internal dimensions. All auxiliary components are training-only, leaving only the compact student at inference.

\subsection{Model Compression and Computational Cost}
\label{subsec:complexity}

At $256\times256$, students retain 0.94--4.6\% of teacher parameters and 1.2--2.5\% of GMACs (Table~\ref{tab:complexity}). Reti-Diff, Restormer, and NAFNet contain 1.19M, 0.46M, and 0.64M restoration parameters; Reti-Diff totals approximately 1.21M with its fixed Retinex module. GFLOPs are approximately twice GMACs. These reductions come from compact architectures; PSD improves their quality without inference overhead. Platform throughput is evaluated in Sec.~\ref{subsec:rov_application}.

\begin{figure*}[!t]
    \centering
    \includegraphics[width=\textwidth]{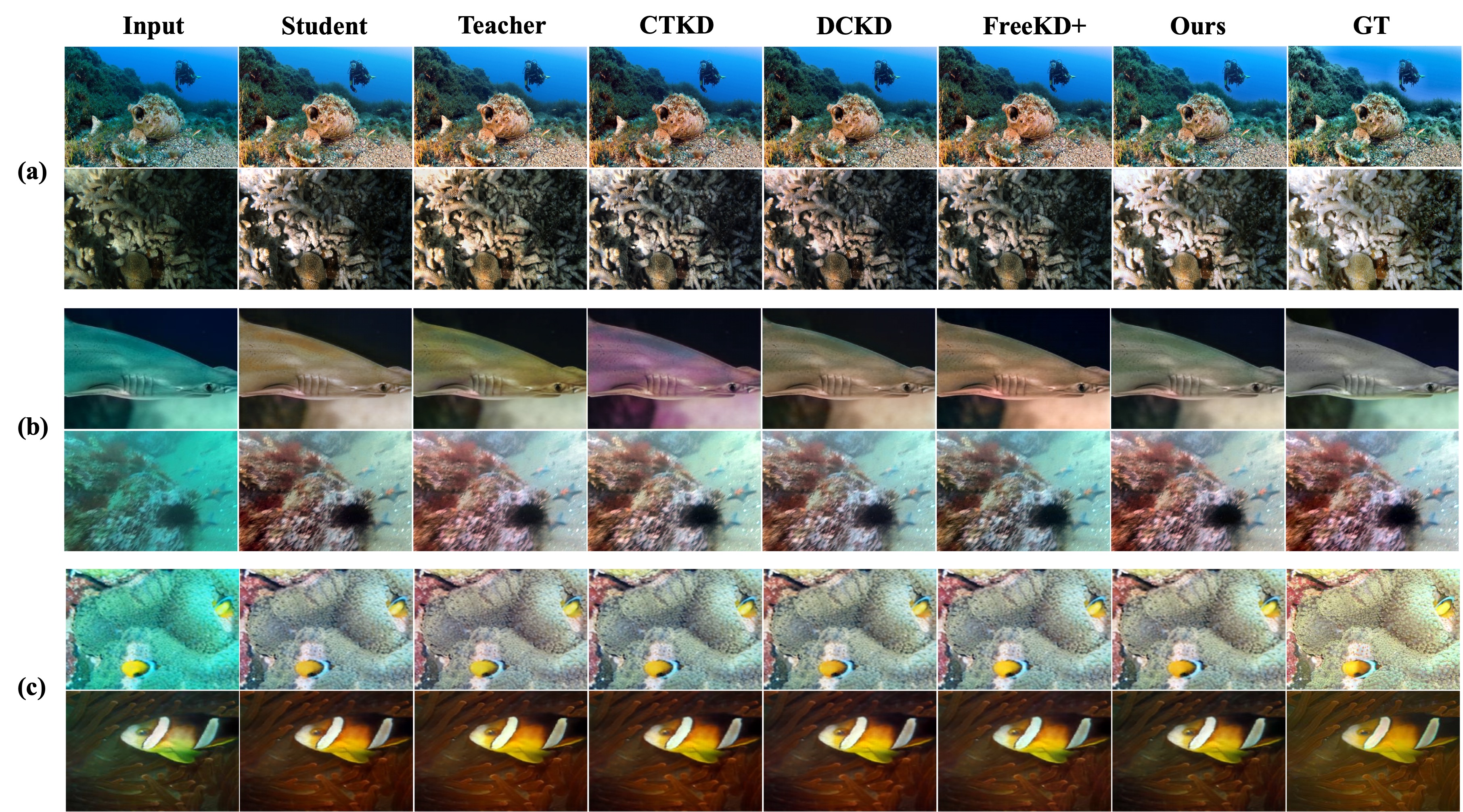}
    \caption{Qualitative comparison for (a) Reti-Diff, (b) Restormer, and (c) NAFNet. Columns show input, task-only student, teacher, CTKD, DCKD, FreeKD+, PSD, and ground truth.}
    \label{fig:qualitative_all}
\end{figure*}

\begin{figure}[!t]
    \centering
    \includegraphics[width=0.92\columnwidth]{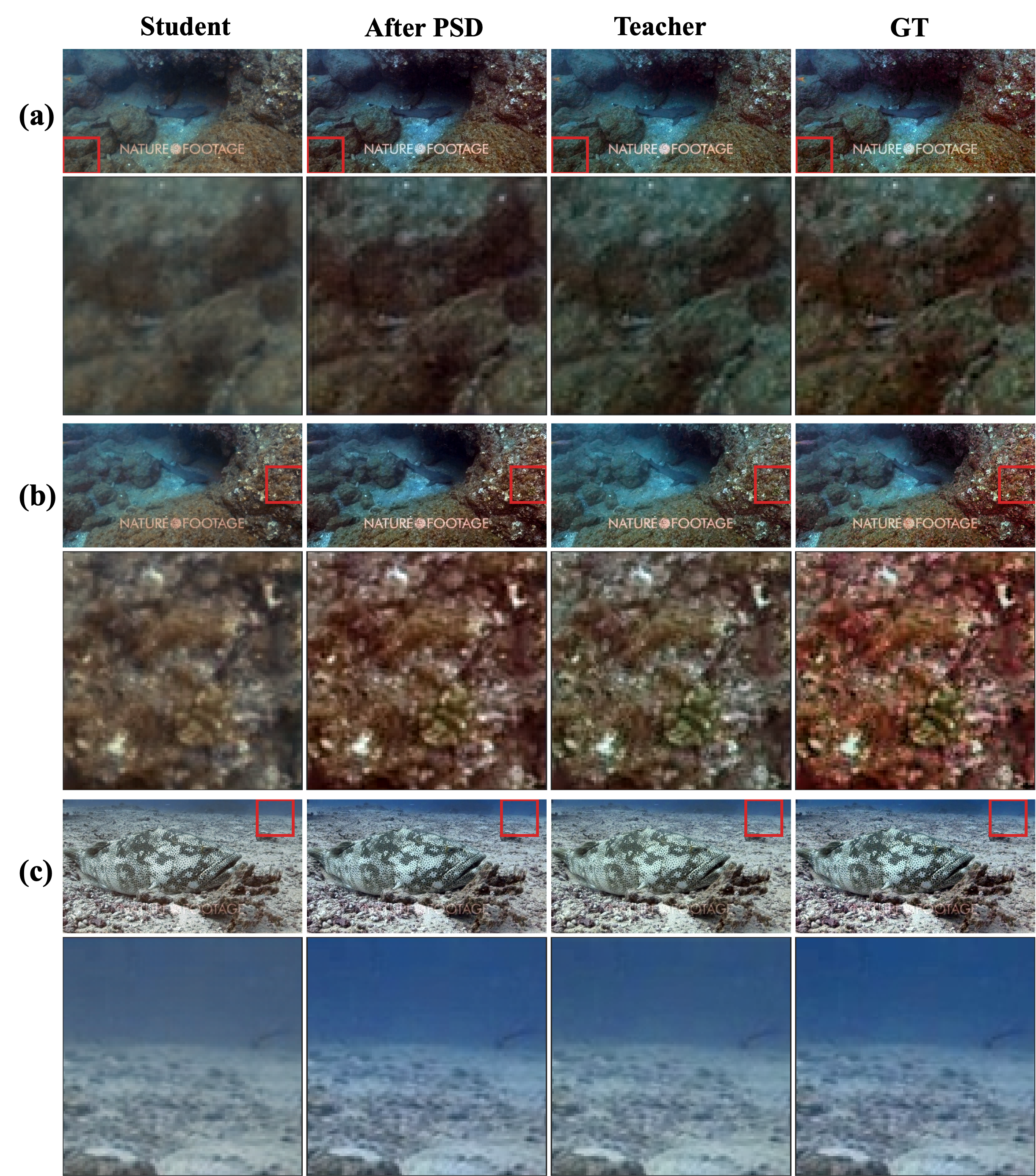}
    \caption{Detail comparison of task-only student, PSD student, teacher, and ground truth. Red-box crops in examples (a) and (b) highlight local contrast, color, and structure.}
    \label{fig:psd_detail}
\end{figure}

\subsection{Comparison with Teachers and Distillation Methods}
\label{subsec:main_results}

Table~\ref{tab:main_results} compares teachers and compact students under matched training budgets. With 95.45\%--99.06\% fewer parameters, PSD students reduced the mean signed teacher--student gap across nine settings from 1.191 to 0.501~dB in PSNR and from 0.0156 to 0.0059 in SSIM. A negative gap denotes a student exceeding its teacher. PSD improved both metrics over task-only training in every setting and led 12 of 18 compact-student comparisons.

For Reti-Diff, the mean PSNR/SSIM gaps decreased from 1.767/0.0068 to 1.350/0.0031. On UIEB, PSD exceeded teacher PSNR by 0.23~dB but retained a 0.0040 SSIM deficit. On LSUI, it narrowed the PSNR deficit from 4.20 to 3.98~dB, whereas other distillation methods enlarged it. Thus, substantial compression losses remain for this backbone. For Restormer, the mean PSNR/SSIM gaps decreased from 1.147/0.0305 to 0.287/0.0108, with the largest recovery on EUVP. PSD led student PSNR on UIEB and LSUI, but CTKD achieved higher SSIM; DCKD marginally led on EUVP. For NAFNet, the mean gaps changed from 0.660/0.0094 to $-0.133$/0.0038, and PSD led all six student comparisons. On LSUI, it exceeded teacher PSNR by 0.44~dB, although SSIM remained 0.0082 lower. 

The ablations below support complementary roles for degradation weighting, spectral transfer, and RGB alignment rather than uniform teacher imitation. Figures~\ref{fig:qualitative_all} and~\ref{fig:psd_detail} illustrate the resulting appearance and detail recovery. A plausible explanation is that band-specific objectives separate appearance correction from structural recovery, while reliability gating suppresses spectral targets that are less accurate than the student's predictions. This selective guidance may reduce conflicting supervision when student capacity is limited.

\subsection{Ablation Study}
\label{subsec:ablation}

We performed leave-one-component-out ablations on LSUI for all three backbone architectures. Table~\ref{tab:lsui_ablation} reports the effects of individually removing the physical head, low- or high-frequency transfer, spectral reliability gating, or image-space compensation. Every removal reduced both metrics, supporting each component's contribution to the complete framework.

\begin{figure*}[!t]
    \centering
    \includegraphics[width=\textwidth]{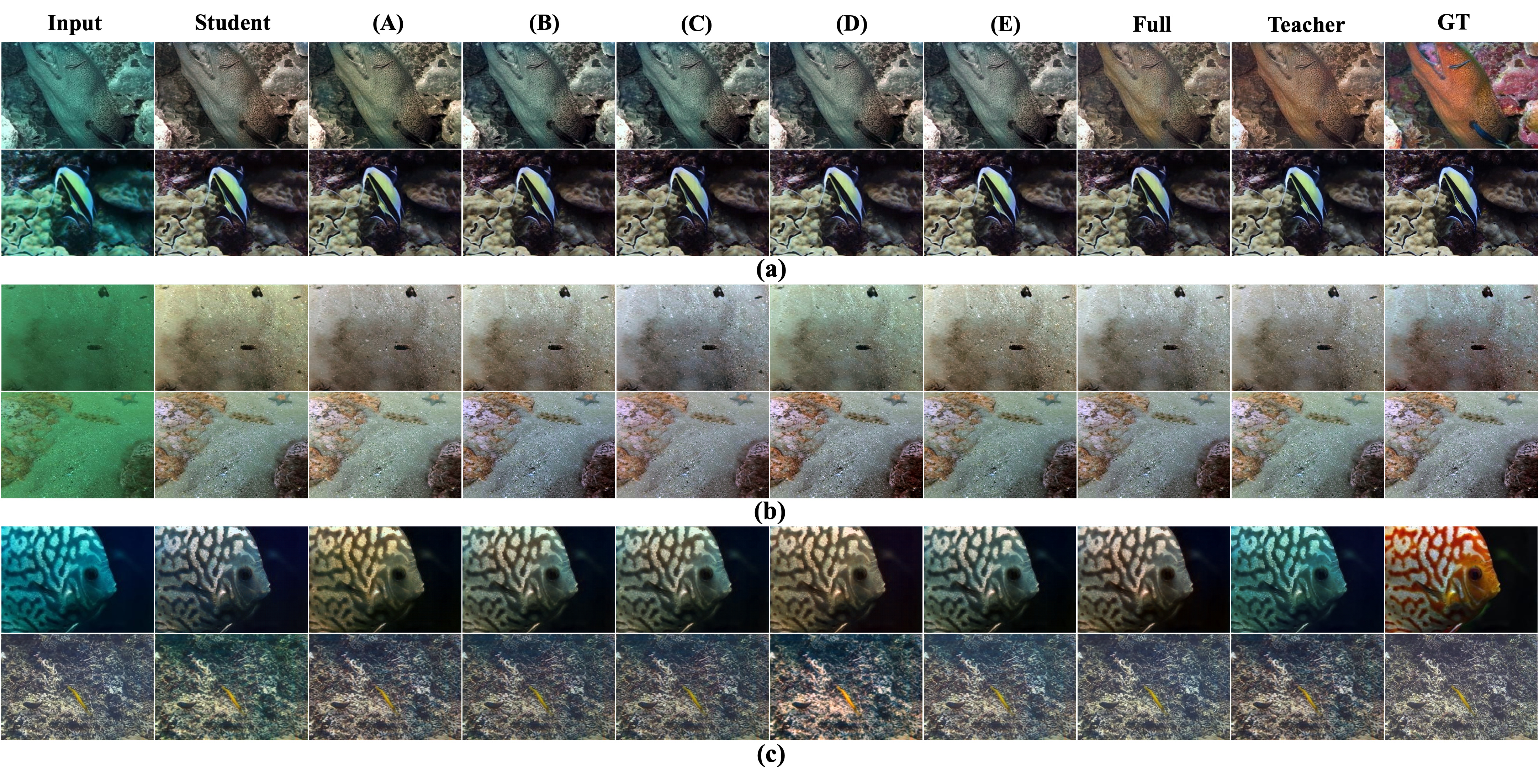}
    \caption{Qualitative ablation of PSD on (a) Reti-Diff, (b) Restormer, and (c) NAFNet. From left to right, (A)--(E) denote w/o PH, w/o LL, w/o HF, w/o Rel., and w/o Img., matching the component order in Table~\ref{tab:lsui_ablation}. \emph{Full} denotes the complete PSD model. Removing individual components produces complementary degradations in color, illumination, contrast, or local structure.}
    \label{fig:ablation_visual}
\end{figure*}

The dominant component varied by backbone. Removing LL transfer caused Reti-Diff's largest PSNR drop (0.15~dB), while removing reliability gating caused its largest SSIM drop (0.0026). Restormer was most sensitive to physical weighting (0.40~dB), and NAFNet to image-space compensation (0.61~dB and 0.0066 SSIM). Removing HF transfer reduced SSIM for every backbone. These results support complementary appearance, structure, and reliability constraints; Fig.~\ref{fig:ablation_visual} illustrates their effects. These sensitivities suggest that transmission weighting directs learning toward attenuated regions, whereas LL and RGB constraints address complementary appearance errors. The SSIM decreases without HF transfer or reliability gating are consistent with losing structural guidance or admitting inaccurate teacher coefficients.

\subsection{Downstream Object Detection}
\label{subsec:udd_detection}

We evaluated downstream detection on UDD's 2,227 images of sea cucumbers, sea urchins, and scallops~\cite{liu2020udd}. Each teacher and PSD student generated an enhanced image domain. Under identical splits and a common protocol, a separate YOLOv9s~\cite{wang2024yolov9} was trained and tested in each enhanced domain and the raw domain, avoiding train--test appearance mismatch.

All six enhanced domains improved both aggregate metrics over Raw (Table~\ref{tab:udd_detection}). Restormer led the teachers at 68.81 $\mathrm{mAP}_{50}$ and 29.74 $\mathrm{mAP}_{50:95}$. Among PSD students, Restormer achieved the largest $\mathrm{mAP}_{50}$ gain (3.22 points), and NAFNet the largest $\mathrm{mAP}_{50:95}$ gain (1.85 points).

\begin{figure*}[!t]
    \centering
    \includegraphics[width=\textwidth]{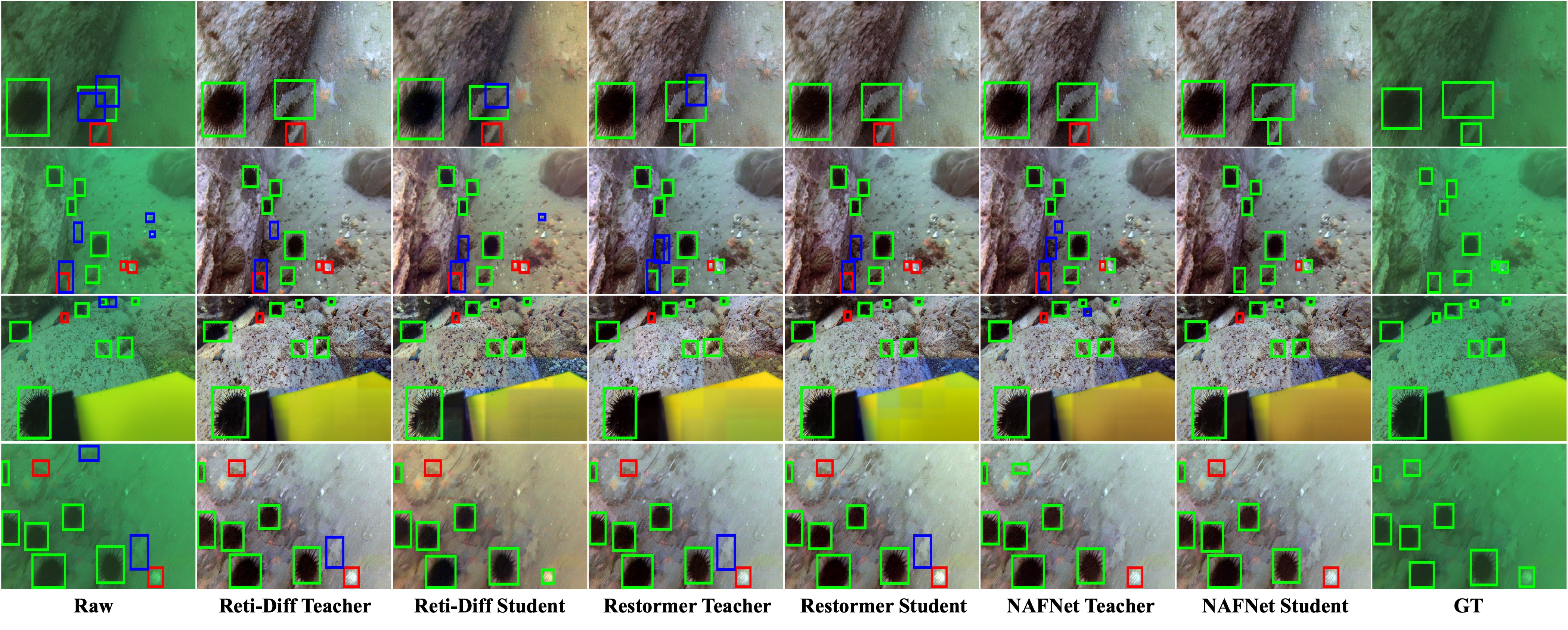}
    \caption{YOLOv9s detections on UDD. Columns: Raw; Reti-Diff teacher/student; Restormer teacher/student; NAFNet teacher/student; ground truth. All students use PSD. Detectors are trained and tested in their corresponding domains.}
    \label{fig:udd_detection}
\end{figure*}

Uneven category gains are consistent with UDD's class imbalance: sea urchins dominate, while sea cucumbers and scallops provide fewer training examples and less stable AP estimates. Enhancement cannot replace this missing supervision, and scallop AP does not improve consistently. Nevertheless, mAP weights classes equally, so aggregate gains are not simply driven by instance counts. Figure~\ref{fig:udd_detection} shows additional organism detections alongside residual misses.

\begin{table*}[!t]
\centering
\caption{Leave-one-component-out ablation on LSUI. PH, LL, HF, Rel., and Img. denote physical weighting, low- and high-frequency transfer, spectral reliability gating, and image-space compensation. All components except the named one are retained; task-only uses none. Arrows indicate performance gains over the task-only baseline.}
\label{tab:lsui_ablation}
\tablefontsize
\setlength{\tabcolsep}{3.5pt}
\renewcommand{\arraystretch}{1.05}
\begin{tabular*}{0.98\textwidth}{@{\extracolsep{\fill}}lcccccc@{}}
\toprule
& \multicolumn{2}{c}{Reti-Diff}
& \multicolumn{2}{c}{Restormer}
& \multicolumn{2}{c}{NAFNet} \\
\cmidrule(lr){2-3}\cmidrule(lr){4-5}\cmidrule(lr){6-7}
Variant
& PSNR $\uparrow$ & SSIM $\uparrow$
& PSNR $\uparrow$ & SSIM $\uparrow$
& PSNR $\uparrow$ & SSIM $\uparrow$ \\
\midrule

Task only
& 24.39
& 0.8680
& 27.79
& 0.8964
& 25.31
& 0.8757 \\

(A) w/o PH
& 24.58 {\scriptsize$\uparrow$0.19}
& 0.8681 {\scriptsize$\uparrow$0.0001}
& 27.78 {\scriptsize$\downarrow$0.01}
& 0.8960 {\scriptsize$\downarrow$0.0004}
& 26.81 {\scriptsize$\uparrow$1.50}
& 0.8840 {\scriptsize$\uparrow$0.0083} \\

(B) w/o LL
& 24.46 {\scriptsize$\uparrow$0.07}
& 0.8679 {\scriptsize$\downarrow$0.0001}
& 27.83 {\scriptsize$\uparrow$0.04}
& 0.8975 {\scriptsize$\uparrow$0.0011}
& 26.81 {\scriptsize$\uparrow$1.50}
& 0.8858 {\scriptsize$\uparrow$0.0101} \\

(C) w/o HF
& 24.49 {\scriptsize$\uparrow$0.10}
& 0.8681 {\scriptsize$\uparrow$0.0001}
& 27.99 {\scriptsize$\uparrow$0.20}
& 0.8978 {\scriptsize$\uparrow$0.0014}
& 27.17 {\scriptsize$\uparrow$1.86}
& 0.8861 {\scriptsize$\uparrow$0.0104} \\

(D) w/o Rel.
& 24.47 {\scriptsize$\uparrow$0.08}
& 0.8668 {\scriptsize$\downarrow$0.0012}
& 27.89 {\scriptsize$\uparrow$0.10}
& 0.8972 {\scriptsize$\uparrow$0.0008}
& 26.78 {\scriptsize$\uparrow$1.47}
& 0.8864 {\scriptsize$\uparrow$0.0107} \\

(E) w/o Img.
& 24.53 {\scriptsize$\uparrow$0.14}
& 0.8676 {\scriptsize$\downarrow$0.0004}
& 28.00 {\scriptsize$\uparrow$0.21}
& 0.8978 {\scriptsize$\uparrow$0.0014}
& 26.67 {\scriptsize$\uparrow$1.36}
& 0.8823 {\scriptsize$\uparrow$0.0066} \\

Full PSD
& \textbf{24.61} {\scriptsize$\uparrow$0.22}
& \textbf{0.8694} {\scriptsize$\uparrow$0.0014}
& \textbf{28.18} {\scriptsize$\uparrow$0.39}
& \textbf{0.8996} {\scriptsize$\uparrow$0.0032}
& \textbf{27.28} {\scriptsize$\uparrow$1.97}
& \textbf{0.8889} {\scriptsize$\uparrow$0.0132} \\

\bottomrule
\end{tabular*}
\end{table*}

\subsection{Application on a Self-Developed ROV}
\label{subsec:rov_application}

\begin{figure}[!t]
    \centering
    \includegraphics[width=0.82\columnwidth]{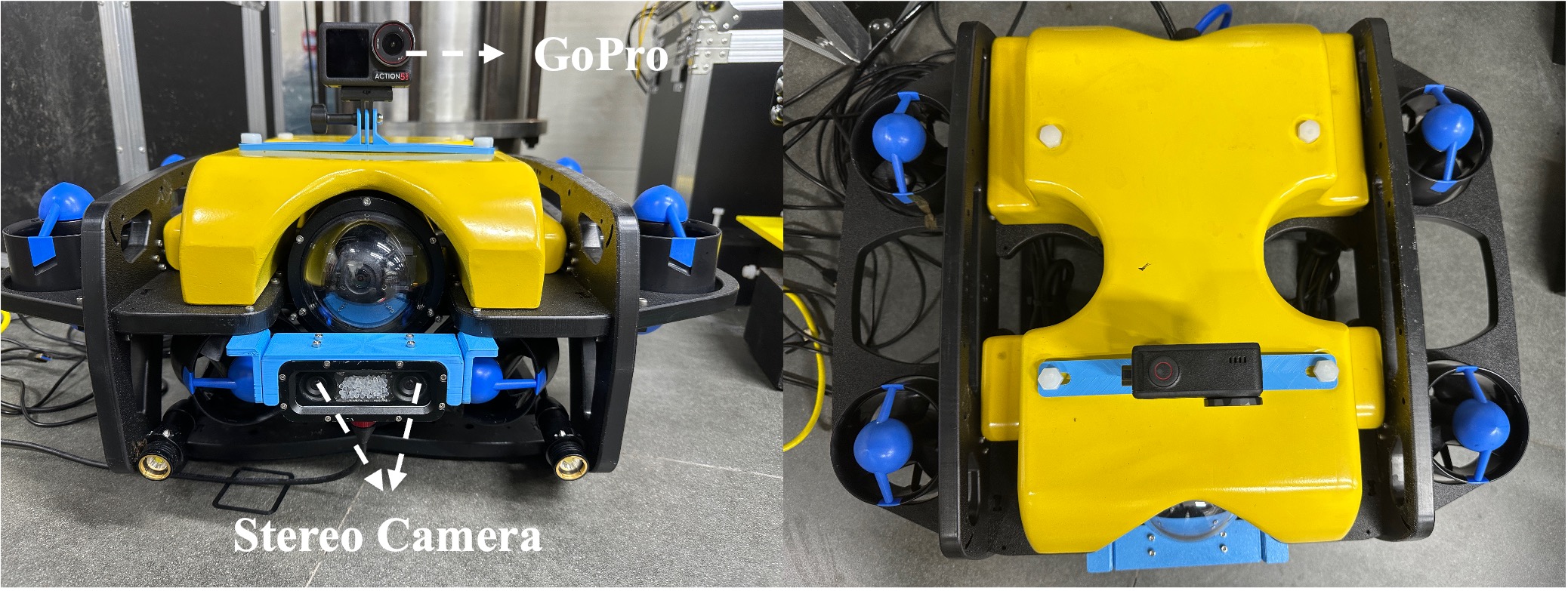}
    \caption{The self-developed ROV used for field data collection in Dushu Lake. Front and top views show the compact vehicle configuration and the locations of the onboard GoPro and stereo camera.}
    \label{fig:rov_platform}
\end{figure}

We deployed all three teacher--student pairs on our self-developed ROV and collected images in Dushu Lake, Suzhou, China. Figure~\ref{fig:rov_platform} shows the vehicle, onboard GoPro, and stereo camera.

At $256\times256$, the students reached 35.7--84.6 FPS, yielding $3.30$--$8.30\times$ teacher speedups (Table~\ref{tab:rov_fps}). All exceeded 30 FPS under this protocol. NAFNet was selected for field deployment because it achieved the highest throughput, despite having more parameters than Restormer.

\begin{table*}[!t]
\centering
\caption{YOLOv9s object detection on UDD. Each detector is trained and tested on the corresponding raw or enhanced image domain. Values are percentages; higher is better. The best result in each column is bold.}
\label{tab:udd_detection}
\tablefontsize
\setlength{\tabcolsep}{3.5pt}
\renewcommand{\arraystretch}{1.12}
\begin{tabular*}{0.98\textwidth}{@{\extracolsep{\fill}}lcccccccc@{}}
\toprule
& \multicolumn{4}{c}{$\mathrm{AP}_{50}$} & \multicolumn{4}{c}{$\mathrm{AP}_{50:95}$} \\
\cmidrule(lr){2-5}\cmidrule(lr){6-9}
Input domain & Sea cucumber & Sea urchin & Scallop & mAP & Sea cucumber & Sea urchin & Scallop & mAP \\
\midrule
Raw                     & 50.18 & 87.56 & 54.46 & 64.06 & 21.04 & 38.01 & 20.85 & 26.63 \\
Reti-Diff teacher       & \textbf{58.80} & 88.97 & 51.66 & 66.47 & 22.85 & 37.14 & 20.43 & 26.80 \\
Reti-Diff student (PSD) & 55.13 & 88.86 & 54.97 & 66.32 & 21.57 & 40.23 & 21.19 & 27.66 \\
Restormer teacher       & 54.90 & \textbf{89.21} & \textbf{62.33} & \textbf{68.81} & 22.74 & \textbf{40.57} & \textbf{25.92} & \textbf{29.74} \\
Restormer student (PSD) & 58.72 & 88.86 & 54.27 & 67.28 & 23.85 & 37.34 & 21.65 & 27.61 \\
NAFNet teacher          & 58.54 & 87.07 & 54.09 & 66.56 & 23.37 & 37.63 & 22.35 & 27.78 \\
NAFNet student (PSD)    & 57.32 & 87.81 & 51.06 & 65.39 & \textbf{24.76} & 39.26 & 21.42 & 28.48 \\
\bottomrule
\end{tabular*}
\end{table*}

\begin{table}[!t]
\centering
\caption{End-to-end inference speed on the self-developed ROV platform using a fixed $256\times256$ input. Higher is better.}
\label{tab:rov_fps}
\tablefontsize
\setlength{\tabcolsep}{3.0pt}
\renewcommand{\arraystretch}{1.15}
\begin{tabular}{lccc}
\toprule
Backbone & Teacher (FPS) & Student (FPS) & Speedup ($\times$) \\
\midrule
Reti-Diff & 5.0  & 36.4          & 7.28 \\
Restormer & 4.3  & 35.7          & \textbf{8.30} \\
NAFNet    & 25.6 & \textbf{84.6} & 3.30 \\
\bottomrule
\end{tabular}
\end{table}

\begin{figure}[!t]
    \centering
    \includegraphics[width=0.92\columnwidth]{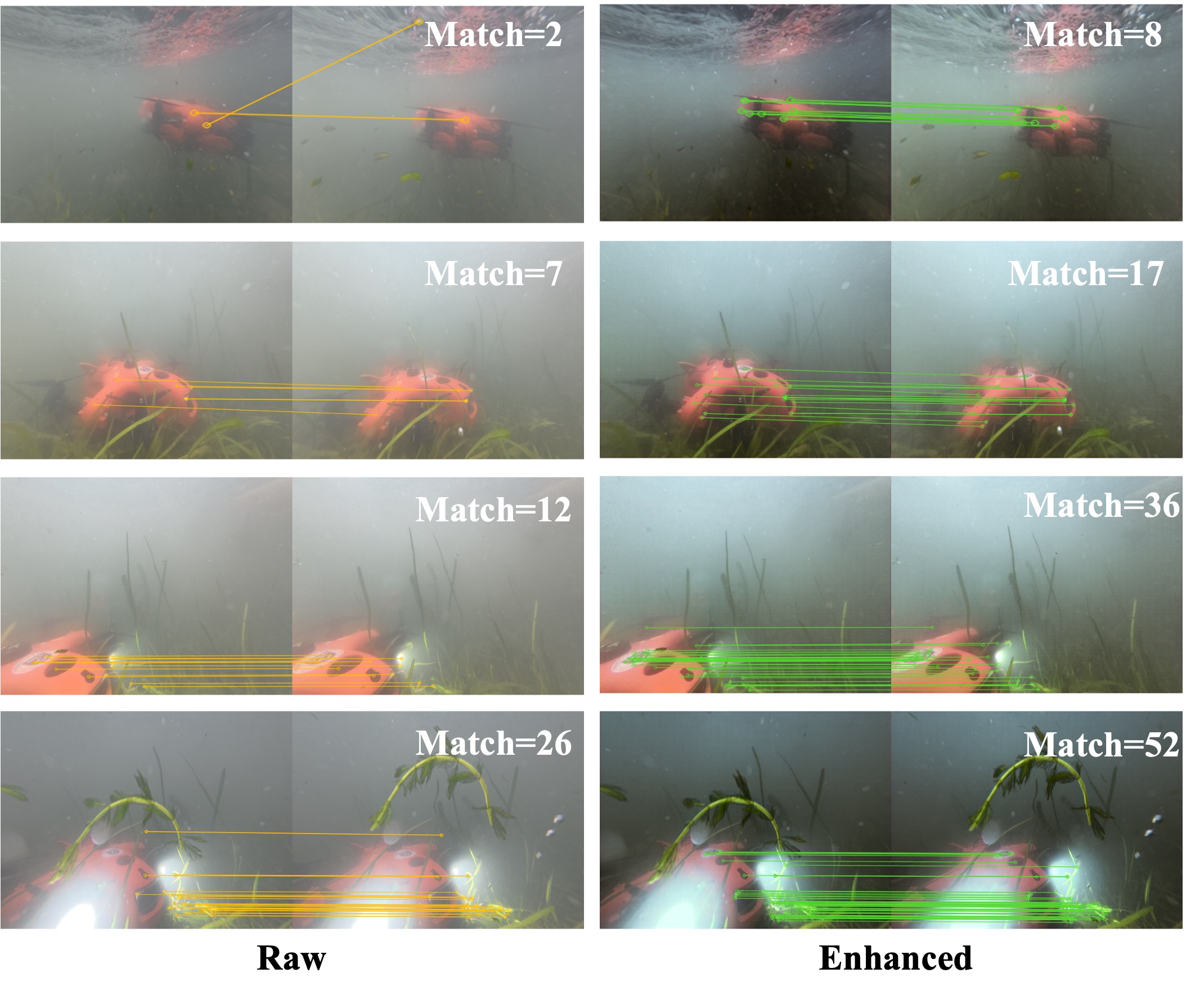}
    \caption{SIFT feature matching on images collected by the ROV. The left half shows matches between raw frame pairs (yellow), and the right half shows matches between the corresponding NAFNet-student-enhanced pairs (green). The numbers below the panels denote the retained matches for each pair.}
    \label{fig:rov_sift}
\end{figure}

The deployed NAFNet student used LSUI-trained weights~\cite{peng2023ushape}. SIFT matches~\cite{lowe2004distinctive} across four frame pairs increased from 2, 7, 12, and 26 to 8, 17, 36, and 52 (Fig.~\ref{fig:rov_sift}), raising the mean from 11.75 to 28.25 ($2.40\times$). More correspondences suggest improved feature visibility but do not establish match correctness or localization accuracy.

On the collected images, UIQM increased by 24.43\% and UCIQE by 29.57\% (Table~\ref{tab:rov_quality}), complementing the feature-matching results. These findings support real-time enhancement on the sampled lake imagery; no-reference scores do not independently establish fidelity to true scene colors.

\begin{table}[!t]
\centering
\caption{No-reference quality scores on the images collected in Dushu Lake. Higher is better.}
\label{tab:rov_quality}
\tablefontsize
\setlength{\tabcolsep}{6pt}
\renewcommand{\arraystretch}{1.15}
\begin{tabular}{lccc}
\toprule
Metric & Raw & Enhanced & Gain (\%) \\
\midrule
UIQM $\uparrow$  & 1.7902 & \textbf{2.2276} & 24.43 \\
UCIQE $\uparrow$ & 0.2184 & \textbf{0.2830} & 29.57 \\
\bottomrule
\end{tabular}
\end{table}

\section{Conclusion}
\label{sec:conclusion}

This paper presented PSD for compressing high-performance UIE models for resource-constrained devices. PSD combines degradation-aware weighting, reliability-gated low- and high-frequency transfer, and image-space compensation without adding inference-time modules. Across three backbones and datasets, PSD consistently improved compact students that retained only 0.94--4.6\% of teacher parameters and 1.2--2.5\% of teacher GMACs. UDD experiments showed benefits for downstream detection, while ROV deployment demonstrated real-time throughput and improved field-image quality. These results support PSD as a practical training framework for efficient underwater enhancement, with long-duration and closed-loop deployment left for future validation.

\bibliographystyle{IEEEtran}
\bibliography{reference}

@article{li2020uieb,
  author  = {Chongyi Li and Chunle Guo and Wenqi Ren and Runmin Cong and Junhui Hou and Sam Kwong and Dacheng Tao},
  title   = {An Underwater Image Enhancement Benchmark Dataset and Beyond},
  journal = {IEEE Transactions on Image Processing},
  volume  = {29},
  pages   = {4376--4389},
  year    = {2020},
  doi     = {10.1109/TIP.2019.2955241}
}

@article{islam2020fast,
  author  = {Md Jahidul Islam and Youya Xia and Junaed Sattar},
  title   = {Fast Underwater Image Enhancement for Improved Visual Perception},
  journal = {IEEE Robotics and Automation Letters},
  volume  = {5},
  number  = {2},
  pages   = {3227--3234},
  year    = {2020},
  doi     = {10.1109/LRA.2020.2974710}
}

@article{li2021ucolor,
  author  = {Chongyi Li and Saeed Anwar and Junhui Hou and Runmin Cong and Chunle Guo and Wenqi Ren},
  title   = {Underwater Image Enhancement via Medium Transmission-Guided Multi-Color Space Embedding},
  journal = {IEEE Transactions on Image Processing},
  volume  = {30},
  pages   = {4985--5000},
  year    = {2021},
  doi     = {10.1109/TIP.2021.3076367}
}

@article{zhuang2022hyperlaplacian,
  author  = {Peixian Zhuang and J. Wu and Fatih Porikli and Chongyi Li},
  title   = {Underwater Image Enhancement With Hyper-Laplacian Reflectance Priors},
  journal = {IEEE Transactions on Image Processing},
  volume  = {31},
  pages   = {5442--5455},
  year    = {2022},
  doi     = {10.1109/TIP.2022.3196546}
}

@article{wang2023tuda,
  author  = {Zhengyong Wang and Liquan Shen and Mai Xu and Mei Yu and Kun Wang and Yufei Lin},
  title   = {Domain Adaptation for Underwater Image Enhancement},
  journal = {IEEE Transactions on Image Processing},
  volume  = {32},
  pages   = {1442--1457},
  year    = {2023},
  doi     = {10.1109/TIP.2023.3244647}
}

@article{peng2023ushape,
  author  = {Lintao Peng and Chunli Zhu and Liheng Bian},
  title   = {U-Shape Transformer for Underwater Image Enhancement},
  journal = {IEEE Transactions on Image Processing},
  volume  = {32},
  pages   = {2593--2607},
  year    = {2023},
  doi     = {10.1109/TIP.2023.3276332}
}

@article{cong2023pugan,
  author  = {Runmin Cong and Wenyu Yang and Wei Zhang and Chongyi Li and Chun-Le Guo and Qingming Huang and Sam Kwong},
  title   = {{PUGAN}: Physical Model-Guided Underwater Image Enhancement Using {GAN} With Dual-Discriminators},
  journal = {IEEE Transactions on Image Processing},
  volume  = {32},
  pages   = {4472--4485},
  year    = {2023},
  doi     = {10.1109/TIP.2023.3286263}
}

@inproceedings{mu2023gupdm,
  author    = {Pan Mu and Hanning Xu and Zheyuan Liu and Zheng Wang and Sixian Chan and Cong Bai},
  title     = {A Generalized Physical-Knowledge-Guided Dynamic Model for Underwater Image Enhancement},
  booktitle = {Proceedings of the ACM International Conference on Multimedia},
  pages     = {7111--7120},
  year      = {2023},
  doi       = {10.1145/3581783.3612323}
}

@inproceedings{zhao2024wfdiff,
  author    = {Chen Zhao and Weiling Cai and Chenyu Dong and Chengwei Hu},
  title     = {Wavelet-Based Fourier Information Interaction With Frequency Diffusion Adjustment for Underwater Image Restoration},
  booktitle = {Proceedings of the IEEE/CVF Conference on Computer Vision and Pattern Recognition},
  pages     = {8281--8291},
  year      = {2024}
}

@article{shi2024cpdm,
  author  = {Xiaowen Shi and Yuan-Gen Wang},
  title   = {{CPDM}: Content-Preserving Diffusion Model for Underwater Image Enhancement},
  journal = {Scientific Reports},
  volume  = {14},
  number  = {1},
  pages   = {31309},
  year    = {2024},
  doi     = {10.1038/s41598-024-82803-y}
}

@inproceedings{he2025retidiff,
  author    = {Chunming He and Chengyu Fang and Yulun Zhang and Longxiang Tang and Jinfa Huang and Kai Li and Zhenhua Guo and Xiu Li and Sina Farsiu},
  title     = {Reti-Diff: Illumination Degradation Image Restoration With Retinex-Based Latent Diffusion Model},
  booktitle = {International Conference on Learning Representations},
  year      = {2025}
}

@inproceedings{peng2025ssuie,
  author    = {Lintao Peng and Liheng Bian},
  title     = {Adaptive Dual-Domain Learning for Underwater Image Enhancement},
  booktitle = {Proceedings of the AAAI Conference on Artificial Intelligence},
  volume    = {39},
  number    = {6},
  pages     = {6461--6469},
  year      = {2025},
  doi       = {10.1609/aaai.v39i6.32692}
}

@inproceedings{hinton2015distilling,
  author    = {Geoffrey Hinton and Oriol Vinyals and Jeff Dean},
  title     = {Distilling the Knowledge in a Neural Network},
  booktitle = {NeurIPS Deep Learning and Representation Learning Workshop},
  year      = {2015}
}

@inproceedings{romero2015fitnets,
  author    = {Adriana Romero and Nicolas Ballas and Samira Ebrahimi Kahou and Antoine Chassang and Carlo Gatta and Yoshua Bengio},
  title     = {{FitNets}: Hints for Thin Deep Nets},
  booktitle = {International Conference on Learning Representations},
  year      = {2015}
}

@inproceedings{zagoruyko2017attention,
  author    = {Sergey Zagoruyko and Nikos Komodakis},
  title     = {Paying More Attention to Attention: Improving the Performance of Convolutional Neural Networks via Attention Transfer},
  booktitle = {International Conference on Learning Representations},
  year      = {2017}
}

@inproceedings{park2019rkd,
  author    = {Wonpyo Park and Dongju Kim and Yan Lu and Minsu Cho},
  title     = {Relational Knowledge Distillation},
  booktitle = {Proceedings of the IEEE/CVF Conference on Computer Vision and Pattern Recognition},
  pages     = {3967--3976},
  year      = {2019}
}

@inproceedings{zamir2022restormer,
  author    = {Syed Waqas Zamir and Aditya Arora and Salman Khan and Munawar Hayat and Fahad Shahbaz Khan and Ming-Hsuan Yang},
  title     = {Restormer: Efficient Transformer for High-Resolution Image Restoration},
  booktitle = {Proceedings of the IEEE/CVF Conference on Computer Vision and Pattern Recognition},
  pages     = {5728--5739},
  year      = {2022}
}

@inproceedings{chen2022nafnet,
  author    = {Liangyu Chen and Xiaojie Chu and Xiangyu Zhang and Jian Sun},
  title     = {Simple Baselines for Image Restoration},
  booktitle = {Proceedings of the European Conference on Computer Vision},
  pages     = {17--33},
  year      = {2022},
  doi       = {10.1007/978-3-031-20071-7_2}
}

@inproceedings{zhang2022wkd,
  author    = {Linfeng Zhang and Xin Chen and Xiaobing Tu and Pengfei Wan and Ning Xu and Kaisheng Ma},
  title     = {Wavelet Knowledge Distillation: Towards Efficient Image-to-Image Translation},
  booktitle = {Proceedings of the IEEE/CVF Conference on Computer Vision and Pattern Recognition},
  pages     = {12464--12474},
  year      = {2022}
}

@inproceedings{li2023ctkd,
  author    = {Zheng Li and Xiang Li and Lingfeng Yang and Borui Zhao and Renjie Song and Lei Luo and Jun Li and Jian Yang},
  title     = {Curriculum Temperature for Knowledge Distillation},
  booktitle = {Proceedings of the AAAI Conference on Artificial Intelligence},
  volume    = {37},
  number    = {2},
  pages     = {1504--1512},
  year      = {2023},
  doi       = {10.1609/aaai.v37i2.25236}
}

@article{xia2023mrda,
  author  = {Bin Xia and Yapeng Tian and Yulun Zhang and Yucheng Hang and Wenming Yang and Qingmin Liao},
  title   = {Meta-Learning-Based Degradation Representation for Blind Super-Resolution},
  journal = {IEEE Transactions on Image Processing},
  volume  = {32},
  pages   = {3383--3396},
  year    = {2023},
  doi     = {10.1109/TIP.2023.3283922}
}

@inproceedings{pham2024fam,
  author    = {Cuong Pham and Van-Anh Nguyen and Trung Le and Dinh Phung and Gustavo Carneiro and Thanh-Toan Do},
  title     = {Frequency Attention for Knowledge Distillation},
  booktitle = {Proceedings of the IEEE/CVF Winter Conference on Applications of Computer Vision},
  pages     = {2277--2286},
  year      = {2024}
}

@inproceedings{zhang2024freekd,
  author    = {Yuan Zhang and Tao Huang and Jiaming Liu and Tao Jiang and Kuan Cheng and Shanghang Zhang},
  title     = {{FreeKD}: Knowledge Distillation via Semantic Frequency Prompt},
  booktitle = {Proceedings of the IEEE/CVF Conference on Computer Vision and Pattern Recognition},
  pages     = {15931--15940},
  year      = {2024}
}

@article{zhang2026freekdplus,
  author  = {Yuan Zhang and Tao Huang and Gaole Dai and Jiaming Liu and Wenzhao Zheng and Jiwen Lu and Shanghang Zhang},
  title   = {{FreeKD+}: A Frequency Knowledge Distillation Framework for Dense Prediction},
  journal = {IEEE Transactions on Pattern Analysis and Machine Intelligence},
  year    = {2026},
  note    = {Early Access},
  doi     = {10.1109/TPAMI.2026.3693362}
}

@inproceedings{zhou2025dckd,
  author    = {Yunshuai Zhou and Junbo Qiao and Jincheng Liao and Wei Li and Simiao Li and Jiao Xie and Yunhang Shen and Jie Hu and Shaohui Lin},
  title     = {Dynamic Contrastive Knowledge Distillation for Efficient Image Restoration},
  booktitle = {Proceedings of the AAAI Conference on Artificial Intelligence},
  volume    = {39},
  number    = {10},
  pages     = {10861--10869},
  year      = {2025},
  doi       = {10.1609/aaai.v39i10.33180}
}

@article{mallat1989wavelet,
  author  = {Stephane G. Mallat},
  title   = {A Theory for Multiresolution Signal Decomposition: The Wavelet Representation},
  journal = {IEEE Transactions on Pattern Analysis and Machine Intelligence},
  volume  = {11},
  number  = {7},
  pages   = {674--693},
  year    = {1989},
  doi     = {10.1109/34.192463}
}

@article{jaffe1990model,
  author  = {Jules S. Jaffe},
  title   = {Computer Modeling and the Design of Optimal Underwater Imaging Systems},
  journal = {IEEE Journal of Oceanic Engineering},
  volume  = {15},
  number  = {2},
  pages   = {101--111},
  year    = {1990},
  doi     = {10.1109/48.50695}
}

@inproceedings{ancuti2012fusion,
  author    = {Cosmin Ancuti and Codruta O. Ancuti and Tom Haber and Philippe Bekaert},
  title     = {Enhancing Underwater Images and Videos by Fusion},
  booktitle = {Proceedings of the IEEE Conference on Computer Vision and Pattern Recognition},
  pages     = {81--88},
  year      = {2012},
  doi       = {10.1109/CVPR.2012.6247661}
}

@inproceedings{drews2013transmission,
  author    = {Paulo Drews Jr. and Erickson R. do Nascimento and Filipe Moraes and Silvia S. C. Botelho and Mario F. M. Campos},
  title     = {Transmission Estimation in Underwater Single Images},
  booktitle = {Proceedings of the IEEE International Conference on Computer Vision Workshops},
  pages     = {825--830},
  year      = {2013},
  doi       = {10.1109/ICCVW.2013.113}
}

@inproceedings{fabbri2018ugan,
  author    = {Cameron Fabbri and Md Jahidul Islam and Junaed Sattar},
  title     = {Enhancing Underwater Imagery Using Generative Adversarial Networks},
  booktitle = {Proceedings of the IEEE International Conference on Robotics and Automation},
  pages     = {7159--7165},
  year      = {2018},
  doi       = {10.1109/ICRA.2018.8460552}
}

@inproceedings{akkaynak2018revised,
  author    = {Derya Akkaynak and Tali Treibitz},
  title     = {A Revised Underwater Image Formation Model},
  booktitle = {Proceedings of the IEEE Conference on Computer Vision and Pattern Recognition},
  pages     = {6723--6732},
  year      = {2018}
}

@inproceedings{akkaynak2019seathru,
  author    = {Derya Akkaynak and Tali Treibitz},
  title     = {{Sea-Thru}: A Method for Removing Water From Underwater Images},
  booktitle = {Proceedings of the IEEE/CVF Conference on Computer Vision and Pattern Recognition},
  pages     = {1682--1691},
  year      = {2019}
}

@article{li2020uwcnn,
  author  = {Chongyi Li and Saeed Anwar and Fatih Porikli},
  title   = {Underwater Scene Prior Inspired Deep Underwater Image and Video Enhancement},
  journal = {Pattern Recognition},
  volume  = {98},
  pages   = {107038},
  year    = {2020},
  doi     = {10.1016/j.patcog.2019.107038}
}

@article{lowe2004distinctive,
  author  = {David G. Lowe},
  title   = {Distinctive Image Features from Scale-Invariant Keypoints},
  journal = {International Journal of Computer Vision},
  volume  = {60},
  number  = {2},
  pages   = {91--110},
  year    = {2004},
  doi     = {10.1023/B:VISI.0000029664.99615.94}
}

@article{liu2020udd,
  author  = {Chongwei Liu and Zhihui Wang and Shijie Wang and Tao Tang and Yulong Tao and Caifei Yang and Haojie Li and Xing Liu and Xin Fan},
  title   = {A New Dataset, Poisson {GAN} and {AquaNet} for Underwater Object Grabbing},
  journal = {arXiv preprint arXiv:2003.01446},
  year    = {2020},
  doi     = {10.48550/arXiv.2003.01446}
}

@article{wang2024yolov9,
  author  = {Chien-Yao Wang and I-Hau Yeh and Hong-Yuan Mark Liao},
  title   = {{YOLOv9}: Learning What You Want to Learn Using Programmable Gradient Information},
  journal = {arXiv preprint arXiv:2402.13616},
  year    = {2024},
  doi     = {10.48550/arXiv.2402.13616}
}

@article{zhang2024liteenhancenet,
  author  = {Song Zhang and Shili Zhao and Dong An and Daoliang Li and Ran Zhao},
  title   = {{LiteEnhanceNet}: A Lightweight Network for Real-Time Single Underwater Image Enhancement},
  journal = {Expert Systems with Applications},
  volume  = {240},
  pages   = {122546},
  year    = {2024},
  doi     = {10.1016/j.eswa.2023.122546}
}

@article{zheng2024mfm,
  author  = {Shijian Zheng and Rujing Wang and Shitao Zheng and Fenmei Wang and Liusan Wang and Zhigui Liu},
  title   = {A Multi-Scale Feature Modulation Network for Efficient Underwater Image Enhancement},
  journal = {Journal of King Saud University--Computer and Information Sciences},
  volume  = {36},
  number  = {1},
  pages   = {101888},
  year    = {2024},
  doi     = {10.1016/j.jksuci.2023.101888}
}

@article{sun2025wateroptical,
  author  = {Zhe Sun and Xuelong Li},
  title   = {Water-Related Optical Imaging: From Algorithm to Hardware},
  journal = {Science China Technological Sciences},
  volume  = {68},
  number  = {1},
  pages   = {1100401},
  year    = {2025},
  doi     = {10.1007/s11431-023-2614-8}
}

@article{sun2026extremedepth,
  author  = {Zhe Sun and Tong Tian and Haofeng Hu and Yan He and Mingjia Shangguan and Tao Yu and Qingsong Yang and Mingliang Chen and Xinwei Wang and Yifan Chen and Kanzhong Yao and Ye Zheng and Ye Qian and Mingyu Dou and Jinghan Xu and Qiang Li and Guojun Wu and Xuelong Li},
  title   = {Extreme-Depth Water-Related Optical Imaging: Conquering Ultra-Low Illumination Environments from Epipelagic Zone to Mariana Trench},
  journal = {PhotoniX},
  volume  = {7},
  pages   = {7},
  year    = {2026},
  doi     = {10.1186/s43074-025-00212-4}
}

@article{chen2023iegi,
  author  = {Yifan Chen and Zhe Sun and Chen Li and Xuelong Li},
  title   = {Computational Ghost Imaging in Turbulent Water Based on Self-Supervised Information Extraction Network},
  journal = {Optics \& Laser Technology},
  volume  = {167},
  pages   = {109735},
  year    = {2023},
  doi     = {10.1016/j.optlastec.2023.109735}
}

@article{chen2025aegi,
  author  = {Yifan Chen and Tong Tian and Xin Lu and Chen Li and Ruolan Zhu and Zhe Sun and Xuelong Li},
  title   = {Attention-Enhanced Computational Ghost Imaging},
  journal = {Science China Information Sciences},
  volume  = {68},
  number  = {6},
  pages   = {162104},
  year    = {2025},
  doi     = {10.1007/s11432-024-4434-5}
}

@article{li2026bridging,
  author  = {Chen Li and Zhe Sun and Xuelong Li},
  title   = {Bridging Physics and Priors: A Unified Diffusion Framework for Underwater Image Restoration},
  journal = {IEEE Transactions on Multimedia},
  year    = {2026},
  note    = {Early Access},
  doi     = {10.1109/TMM.2026.3697673}
}

@article{jiang2024pdd,
  author  = {Qiuping Jiang and Yaozu Kang and Zhihua Wang and Wenqi Ren and Chongyi Li},
  title   = {Perception-Driven Deep Underwater Image Enhancement Without Paired Supervision},
  journal = {IEEE Transactions on Multimedia},
  volume  = {26},
  pages   = {4884--4897},
  year    = {2024},
  doi     = {10.1109/TMM.2023.3327613}
}

@article{zhou2024pdr,
  author  = {Jingchun Zhou and Shiyin Wang and Zifan Lin and Qiuping Jiang and Ferdous Sohel},
  title   = {A Pixel Distribution Remapping and Multi-Prior Retinex Variational Model for Underwater Image Enhancement},
  journal = {IEEE Transactions on Multimedia},
  volume  = {26},
  pages   = {7838--7849},
  year    = {2024},
  doi     = {10.1109/TMM.2024.3372400}
}

\newpage

\end{document}